\documentclass[]{spectral_ai}

\usepackage[toc,page,header]{appendix}

\usepackage{wrapfig}
\usepackage{multirow}
\usepackage{makecell}
\usepackage{colortbl}
\usepackage{amsfonts}
\usepackage{amsmath}
\usepackage{float}
\usepackage{siunitx}
\usepackage{natbib}
\usepackage{tabularx}
\usepackage{array}
\usepackage{xurl}

\definecolor{colorfirst}{rgb}{.866,.945, 0.831} %
\definecolor{colorsecond}{rgb}{1, 0.98, 0.83} %
\definecolor{colorthird}{rgb}{0.76, 0.87, 0.92} %
\definecolor{colorcite}{rgb}{0.212, 0.490, 0.741}

\title{Simulation-to-real transfer learning for infrared spectroscopic chemical sensing and analysis from molecules to complex samples}

\author[1]{Yusen Tan}
\author[2]{Yixuan Chen}
\author[1]{Zheng Fang}
\author[1]{Pan Liu}
\author[1]{Yifan Li}
\author[3]{Qinyu Guo}
\author[3]{Zhedong Lin}
\author[4]{Yuqiang Li}
\author[5]{Xiangxiang Zeng$^{\dagger}$}
\author[6]{Tong Wang$^{\dagger}$}
\author[1, 7]{Jun Xia$^{\dagger}$}

\affiliation[1]{Information Hub, The Hong Kong University of Science and Technology (Guangzhou), Guangzhou 511453, China} 
\affiliation[2]{College of Computer Science and Technology, Jilin University, Changchun 130012, China}
\affiliation[3]{School of Computer Science, University of Auckland, Auckland 1142, New Zealand}
\affiliation[4]{AI for Chemistry Center, Shanghai Artificial Intelligence Laboratory, Shanghai 200232, China}
\affiliation[5]{College of Computer Science and Electronic Engineering, Hunan University, Changsha 410082, China} 
\affiliation[6]{State Key Laboratory of Chemo and Biosensing, College of Chemistry and Chemical Engineering, Hunan University, Changsha 410082, China}
\affiliation[7]{Department of Computer Science and Engineering, The Hong Kong University of Science and Technology, Hong Kong SAR, China}

\abstract{
Infrared (IR) spectroscopy is widely used for chemical sensing, from molecular characterization to the analysis of complex samples, but extracting reliable chemical information from spectra remains challenging. Conventional interpretation based on expert peak assignment, rule-based analysis, and library matching is labor-intensive and difficult to scale across analytical objectives and datasets while maintaining consistently high accuracy, and its reliance on prior knowledge and reference spectra makes it less reliable for unfamiliar compounds and complex samples. Machine-learning methods have enabled more automated and scalable chemical inference from IR spectra, but most approaches remain tailored to individual tasks or datasets, require large training sets of labeled spectra, and show limited transferability across analytical objectives and experimental datasets. Together, these limitations and the scarcity of labeled experimental IR spectra motivate simulation-to-real transfer learning to enable data-efficient chemical inference under limited supervision. Here we introduce UltraIR, a foundation model for IR spectroscopy with more than 100 million parameters that enables simulation-to-real transfer learning for chemical sensing and analysis from molecules to complex samples. UltraIR learns a shared spectral representation by pretraining on approximately 60 million simulated IR spectra using three complementary pretraining objectives for spectral reconstruction, molecular fingerprint similarity alignment, and functional-group prediction. The pretrained encoder is then adapted to each downstream objective using task-specific labels or targets paired with the corresponding IR spectra, typically with only a limited number of labeled spectra available. Across benchmark evaluations of functional-group prediction, molecular structure elucidation, physicochemical property prediction, and mixture-component identification and quantification, as well as real-world chemical sensing applications involving bacterial classification, medicinal-herb geographic origin traceability and constituent quantification using two newly generated in-house datasets, microplastics classification, and soil property prediction, UltraIR outperforms conventional machine-learning and task-specific deep-learning baselines. UltraIR demonstrates the practical value of simulation-to-real transfer learning through strong performance in two challenging real-world chemical sensing scenarios: downstream adaptation using limited labeled experimental spectra and zero-shot inference for the same analytical task across Fourier-transform infrared spectrometers and laboratories. More broadly, this approach provides a route to adaptable, data-efficient chemical sensing systems for extracting reliable chemical information from IR spectra of complex real-world samples.
}

\correspondence{Xiangxiang Zeng (\href{mailto:xzeng@foxmail.com}{xzeng@foxmail.com}); Tong Wang (\href{mailto:xwangtong@hnu.edu.cn}{wangtong@hnu.edu.cn}); Jun Xia (\href{mailto:junxia@hkust-gz.edu.cn}{junxia@hkust-gz.edu.cn})}

\begin{document}

\newtheorem{example}{Example}
\newtheorem{definition}{Definition}
\newtheorem{lemma}{Lemma}
\newtheorem{theorem}{Theorem}
\newtheorem{proposition}{Proposition}
\newtheorem{corollary}{Corollary}[proposition]
\newtheorem{assumption}{Assumption}
\newtheorem{observation}{Observation}

\newcommand{\fig}[1]{Fig.~\ref{#1}}
\newcommand{\eq}[1]{Eq.~(\ref{#1})}
\newcommand{\tb}[1]{Tab.~\ref{#1}}
\newcommand{\se}[1]{Section~\ref{#1}}
\newcommand{\ap}[1]{Appendix~\ref{#1}}

\newcommand*{\dif}{\mathop{}\!\mathrm{d}}
\newcommand{\bbE}{\ensuremath{\mathbb{E}}}
\newcommand{\bbR}{\ensuremath{\mathbb{R}}}
\newcommand{\caL}{\ensuremath{\mathcal{L}}}
\newcommand{\caD}{\ensuremath{\mathcal{D}}}


\maketitle

\section{Introduction}
\label{sec:intro}

Infrared (IR) spectroscopy is an analytical technique that probes molecular vibrations through their interaction with infrared radiation to generate chemically informative spectral fingerprints \cite{stuart2004infrared, griffiths1983fourier, kazarian2013atr}. Its rapid, label-free, and often non-destructive measurements support widespread use in chemical sensing and analysis, ranging from molecular characterization to complex-sample analysis in environmental monitoring, clinical analysis, and materials characterization \cite{application2, application1, ossl, microp}. However, the ability to acquire IR spectra at scale has not been matched by an equally scalable ability to convert them into reliable chemical information \cite{difficult1, baker2014using, gautam2015review}. Conventional workflows still rely heavily on expert peak assignment, rule-based interpretation, and library matching. These approaches depend on prior expert knowledge, predefined rules, or reference spectra, which can limit reliable interpretation when suitable reference information is unavailable, as may occur for unfamiliar compounds and complex samples \cite{difficult1, houthuijs2023silico}. Such dependencies also make conventional workflows difficult to scale across analytical objectives and datasets while maintaining consistent accuracy \cite{expert1, expert2, stuart2004infrared}. A central challenge for next-generation IR chemical sensing is therefore to translate spectra into reliable chemical information at scale across analytical objectives and datasets while maintaining high accuracy.

Machine learning has begun to address the challenge of automated IR analysis at scale by learning predictive relationships directly from spectral data \cite{lansford2020infrared,ren2022wavelength,zhu2023triboelectric}. Recent studies have shown that IR spectra support functional-group recognition, molecular structure elucidation, library-based molecular retrieval, and mixture-component identification \cite{fg1, fg2, irtomol, denovo, retrieval, deepmir}. Together, these advances demonstrate that IR spectra contain molecular information that can support diverse analytical tasks. However, most approaches are tailored to individual tasks or datasets and require large task-specific collections of labeled spectra, limiting the development of shared spectral representations that transfer across analytical objectives and experimental datasets. Despite enabling more rapid and scalable analysis than expert peak assignment, rule-based interpretation, and library matching, these models often achieve only modest accuracy gains over strong conventional machine-learning baselines, and these gains may disappear when few labeled spectra are available for training. These limitations motivate a general representation-learning approach that enables scalable, reliable, transferable, and data-efficient IR analysis.

Foundation models provide such a general representation-learning approach, as demonstrated by large language models (LLMs), whose large-scale pretraining supports adaptation across diverse downstream tasks \cite{brown2020language,bommasani2021opportunities}. Similar large-scale representation-learning paradigms have emerged across sensing domains, including medical imaging \cite{retfound,cxr_foundation}, continuous glucose monitoring \cite{cgm_foundation}, neural-activity modeling \cite{neural_foundation}, remote sensing \cite{geo}, and robotic tactile sensing \cite{supertac,sun2026spike}. However, foundation-model approaches remain underexplored in IR spectroscopy, with their development constrained by the scarcity of large, chemically diverse collections of labeled experimental spectra \cite{fg2,ibm}. Simulated IR spectra offer a scalable, complementary source of pretraining data that can broaden molecular coverage and support shared representation learning \cite{chemprop,nipsdata,qm9s,USPTO}. To date, however, simulated IR spectra have mainly been used within individual chemical sensing and analysis tasks, with reported performance gains demonstrating their value for data-driven IR analysis \cite{irtomol}. Together, these observations motivate a simulation-to-real representation-learning paradigm that combines large-scale pretraining on simulated spectra with supervised adaptation using task-specific labeled experimental spectra, offering a promising route toward transferable and data-efficient IR foundation models. Specifically, pretraining can capture recurring relationships among IR spectral patterns, molecular structures and chemical properties across a broad chemical space, while downstream supervision adapts the shared representation to each chemical sensing and analysis objective.

Here we introduce UltraIR, a foundation model for IR spectroscopy with more than 100 million parameters that enables simulation-to-real transfer learning for chemical sensing and analysis from molecules to complex samples. UltraIR uses a two-stage learning framework in which a shared spectral encoder is first pretrained on approximately 60 million simulated IR spectra. For each downstream objective, the pretrained encoder is paired with a newly initialized task-specific module, and all components are jointly optimized using labeled experimental IR spectra. The encoder combines a derivative-aware multi-channel input, hierarchical convolutional modules, and a patch-based Transformer. These components capture local line shapes and peak shifts, extract multiscale vibrational signatures, and model longer-range spectral dependencies, respectively. Pretraining integrates three complementary objectives. Wavelet-domain spectral reconstruction encourages preservation of broad spectral envelopes and fine absorption features. Molecular fingerprint similarity alignment organizes the latent space according to graded structural relationships among molecules, while multi-label functional-group prediction anchors the representation to interpretable chemical motifs. Together, this design couples broad molecular exposure from simulated spectra with supervised adaptation using task-specific labeled experimental IR spectra, often available only in limited numbers, to provide a shared representation backbone for molecular interpretation, mixture analysis, and chemical sensing in complex samples.

We evaluate UltraIR across molecular and mixture benchmarks, as well as real-world chemical sensing applications. The benchmarks cover functional-group prediction, molecular structure elucidation, physicochemical property prediction, and mixture-component identification and quantification. The real-world applications include bacterial classification, medicinal-herb geographic origin traceability and constituent quantification using two newly generated in-house datasets, microplastics classification, and soil property prediction. Across these evaluations, UltraIR outperforms conventional machine-learning and task-specific deep-learning baselines. The practical value of UltraIR's simulation-to-real transfer learning is demonstrated under two demanding conditions encountered in real-world chemical sensing: downstream adaptation with limited labeled experimental spectra and zero-shot inference for the same analytical task across Fourier-transform infrared (FTIR) spectrometers and laboratories. More broadly, this learning framework provides a basis for adaptable and data-efficient chemical sensing systems that extract reliable chemical information from IR spectra of complex real-world samples.

\section{Results}
\label{sec:results}

\subsection*{UltraIR overview}

UltraIR is a foundation model for IR spectroscopy with more than 100 million parameters that enables simulation-to-real transfer learning for chemical sensing and analysis from molecules to complex samples. The framework integrates two complementary resources: approximately 60 million simulated IR spectra for large-scale pretraining and a diverse collection of downstream datasets for supervised adaptation and evaluation across molecular-level benchmarks, mixture analysis, and real-world chemical sensing applications (Fig.~\ref{fig:task_landscape}a).

\begin{figure}[p]
    \centering
    \includegraphics[width=0.95\textwidth]{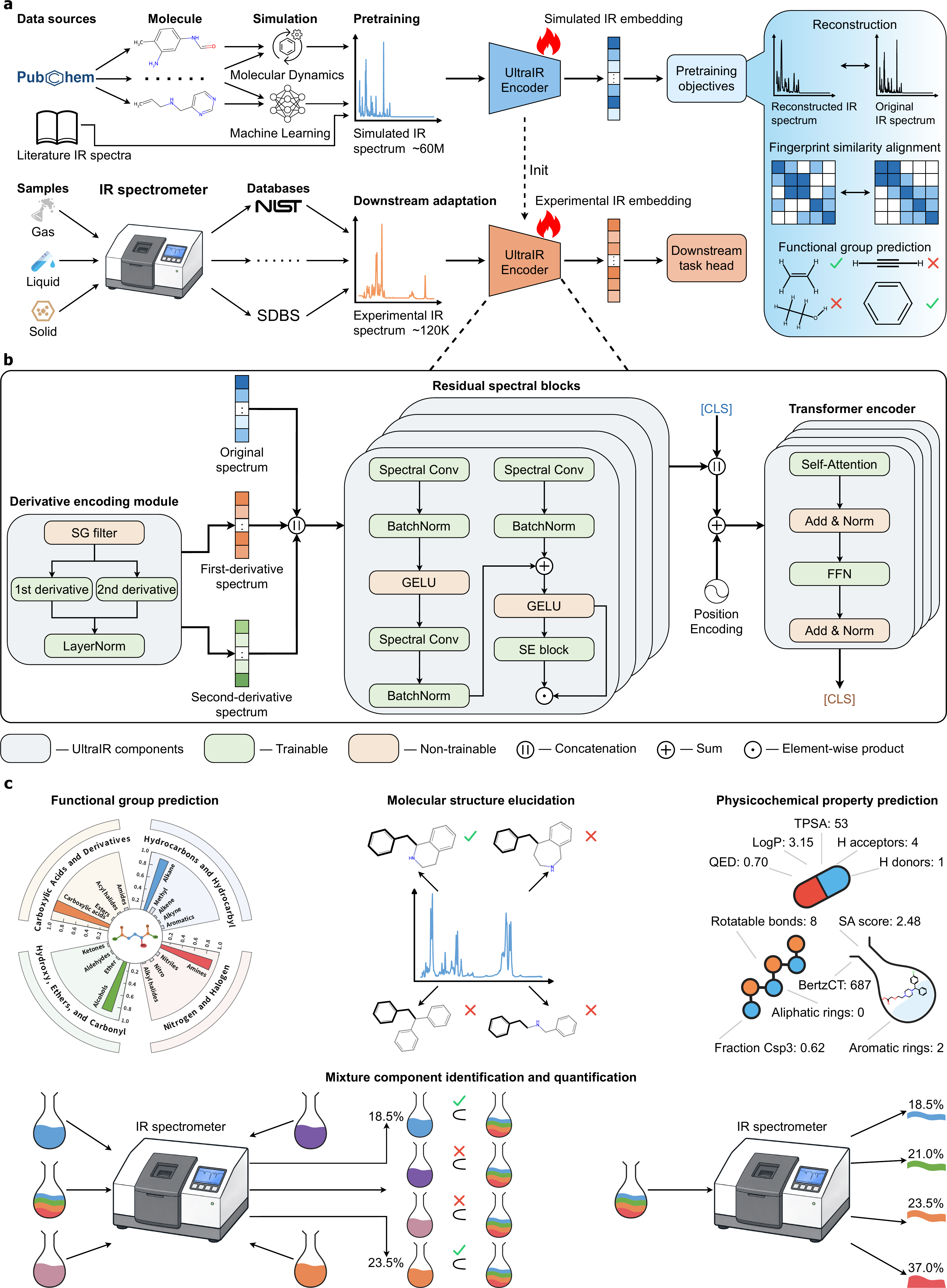}
\end{figure}

The standardized pretraining dataset comprises molecule-associated IR spectra from three sources: approximately 1.5 million publicly released simulated spectra from IRtoMol, the multimodal spectroscopy dataset, and QM9S \citep{irtomol,nipsdata,qm9s}; approximately 7.5 million newly generated spectra from molecular-dynamics simulations; and approximately 51 million spectra generated using a machine-learning-based spectral predictor \citep{chemprop}. Before pretraining, simulated spectra corresponding to molecules present in the downstream molecular datasets were excluded from the pretraining pool to prevent molecule-level overlap. Together, these sources provide the scale and molecular diversity required for large-scale pretraining.

For supervised adaptation and evaluation, we assembled approximately 120,000 experimental IR spectra together with the simulated USPTO molecular benchmark. Molecular-level benchmarks use spectra from the National Institute of Standards and Technology (NIST) Chemistry WebBook \citep{nistwebbook}, the Spectral Database for Organic Compounds (SDBS) \citep{sdbs}, and USPTO \citep{USPTO} for functional-group prediction, molecular structure elucidation, and physicochemical property prediction. Mixture analysis uses NIST, SDBS, and USPTO spectra for targeted component detection and targeted fractional contribution estimation, the DeepMIR liquid-mixture dataset for additional evaluation of targeted component detection \citep{deepmir}, and FTIR mixture spectra for mixture-level component quantification \citep{digital_discovery_2022}. Real-world chemical sensing applications include genus-level classification of bacterial isolates from green snow \citep{bacteria}, medicinal-herb geographic origin traceability and constituent quantification, microplastics classification using environmentally sourced spectra \citep{microp}, and soil property prediction using the Open Soil Spectral Library \citep{ossl}. Together, these datasets span chemical sensing and analysis from molecules and mixtures to biological and botanical samples and complex environmental samples.

UltraIR follows a two-stage simulation-to-real learning framework (Fig.~\ref{fig:task_landscape}a). During pretraining, a shared encoder learns spectral representations from simulated IR spectra through three complementary objectives. Wavelet-domain spectral reconstruction encourages the encoder to retain coarse spectral envelopes and fine absorption features; molecular fingerprint similarity alignment encourages the latent space to reflect graded structural similarity among molecules; and multi-label functional-group prediction links the learned representations to interpretable chemical motifs. Together, these objectives integrate spectral morphology, molecular structural relationships, and functional-group information into a shared representation. For downstream adaptation, the pretrained encoder initializes the shared representation backbone, while a newly initialized task-specific head is added for each downstream objective. The encoder and head are then jointly optimized using the corresponding supervised data. This framework combines broad exposure to chemical diversity through pretraining on simulated spectra with task-specific supervision, enabling transfer of the shared representation across analytical objectives.

\begingroup
\captionsetup{type=figure}
\captionof{figure}{
\textbf{Overview of the UltraIR simulation-to-real framework, encoder architecture, and molecular- and mixture-level analytical tasks.}
\textbf{a,} UltraIR data resources and the two-stage simulation-to-real learning framework. The pretraining dataset contains approximately 60 million simulated IR spectra assembled from publicly released spectrum--molecule pairs and spectra predicted or simulated for PubChem-derived molecules using machine-learning models and molecular dynamics. The shared encoder is pretrained with three chemically motivated objectives: wavelet-domain spectral reconstruction, molecular fingerprint similarity alignment and multi-label functional-group prediction. For downstream adaptation, the pretrained encoder initializes the shared representation backbone and is jointly optimized with a newly initialized task-specific head using supervised downstream data, including approximately 120,000 experimental IR spectra from gas-, liquid-, and solid-phase samples.
\textbf{b,} UltraIR encoder architecture. The original spectrum is concatenated with first- and second-derivative channels generated by the derivative encoding module and processed through hierarchical residual spectral blocks. The resulting features are concatenated with a \texttt{[CLS]} token, combined with positional encodings, and processed by a Transformer encoder to produce the final \texttt{[CLS]} representation. Colors distinguish UltraIR components and separate trainable from non-trainable operations; symbols denote concatenation, summation, and element-wise multiplication.
\textbf{c,} Molecular- and mixture-level analytical tasks used to evaluate UltraIR, comprising functional-group prediction, molecular structure elucidation, physicochemical property prediction, and mixture-component identification and quantification. Mixture analysis includes targeted component detection, targeted fractional contribution estimation, and mixture-level component quantification.
}
\label{fig:task_landscape}
\endgroup

\subsection*{Downstream task landscape}

To evaluate whether the shared spectral representation transfers across analytical objectives and datasets, we assessed UltraIR across a downstream task landscape spanning molecular- and mixture-level benchmarks and real-world chemical sensing applications. The molecular- and mixture-level benchmarks link IR spectra to functional-group annotations, molecular structures, physicochemical properties, and mixture composition (Fig.~\ref{fig:task_landscape}c), whereas the real-world applications evaluate chemical inference from complex biological, botanical, and environmental samples.

Functional-group prediction evaluates whether UltraIR can identify functional groups from their characteristic vibrational features. Molecular structure elucidation evaluates whether UltraIR can identify the correct molecular structure from spectral information under a specified molecular-formula constraint. Physicochemical property prediction examines whether the shared spectral representation encodes continuous molecular attributes beyond discrete structural annotations. Mixture analysis assesses component detection and quantification through three tasks. In targeted component detection, the model receives a single-compound reference spectrum and a mixture spectrum and predicts whether the corresponding compound is present in the mixture. In targeted fractional contribution estimation, the same pair of spectra is used to estimate the target compound's fractional contribution. In mixture-level component quantification, the model receives only a mixture spectrum and predicts the proportions of multiple components.

Real-world chemical sensing applications span four sample domains and multiple analytical objectives. Genus-level bacterial classification evaluates whether biological spectral fingerprints can distinguish fast-growing bacterial isolates collected from green snow. Medicinal-herb characterization combines geographic origin traceability with chemical constituent quantification, requiring both categorical and continuous inference from botanical spectra. Microplastics classification uses environmental IR spectra to distinguish polymer types. Soil property prediction evaluates whether IR spectra support the prediction of continuous soil physicochemical properties.

The following sections benchmark UltraIR against established conventional machine-learning and task-specific deep-learning methods appropriate to each task.

\subsection*{UltraIR enables molecular structural interpretation from infrared spectra}

We first evaluated UltraIR on functional-group prediction and molecular structure elucidation using the public experimental NIST and SDBS datasets and the simulated USPTO benchmark \citep{nistwebbook,sdbs,USPTO}.

\subsubsection*{Functional-group prediction.}

For functional-group prediction, UltraIR was benchmarked against task-specific deep-learning models, including FCGFormer and IRAnalysis \citep{fg1, fg2}, and conventional machine-learning baselines, including XGBoost, random forest (RF), k-nearest neighbors (KNN), and a logistic classifier implemented using scikit-learn \citep{chen2016xgboost, breiman2001random, pedregosa2011scikit}. Performance was evaluated using three standard and complementary metrics: Micro-F1, Macro-F1, and exact match ratio (EMR). Micro-F1 and Macro-F1 are commonly used label-level metrics for multi-label classification. Micro-F1 aggregates true positives, false positives, and false negatives across all functional-group labels, providing an overall measure of label-wise prediction performance, whereas Macro-F1 first computes the F1 score for each functional group and then averages across groups, giving equal weight to frequent and infrequent functional groups. In contrast, EMR provides a molecule-level assessment of complete functional-group assignment: a prediction is counted as correct only when the full set of functional groups is recovered without missing labels or false assignments. We assessed model behavior from three perspectives: overall performance on each dataset, performance variation as a function of labeled training-data size, and performance for individual functional groups.

Across the three datasets and all three metrics, UltraIR consistently and substantially outperformed both task-specific deep-learning models and conventional machine-learning baselines (Fig.~\ref{fig:molecular_interpretation}a). On SDBS and USPTO, FCGFormer and IRAnalysis either only marginally outperformed XGBoost or performed comparably to it. On NIST, both models slightly underperformed XGBoost across all three metrics. These results demonstrate that task-specific deep learning alone does not confer a decisive advantage over strong conventional baselines for functional-group prediction. By contrast, UltraIR established a clear and consistent performance advantage over both classes of methods, with the most pronounced gains in EMR on NIST and USPTO. These gains are chemically meaningful rather than merely statistical. Functional-group prediction converts an IR spectrum into an interpretable molecular annotation that chemists can use to assess molecular composition, validate spectral interpretations, and evaluate plausible structural hypotheses. A partially correct label-wise prediction may still miss a key group or introduce a false assignment, thereby altering the chemical interpretation of the molecule even when Micro-F1 or Macro-F1 remains high. Higher EMR therefore shows that UltraIR more frequently recovers the complete functional-group set, rather than merely improving isolated label detections.

\begin{figure}[p]
    \centering
    \includegraphics[width=0.95\textwidth]{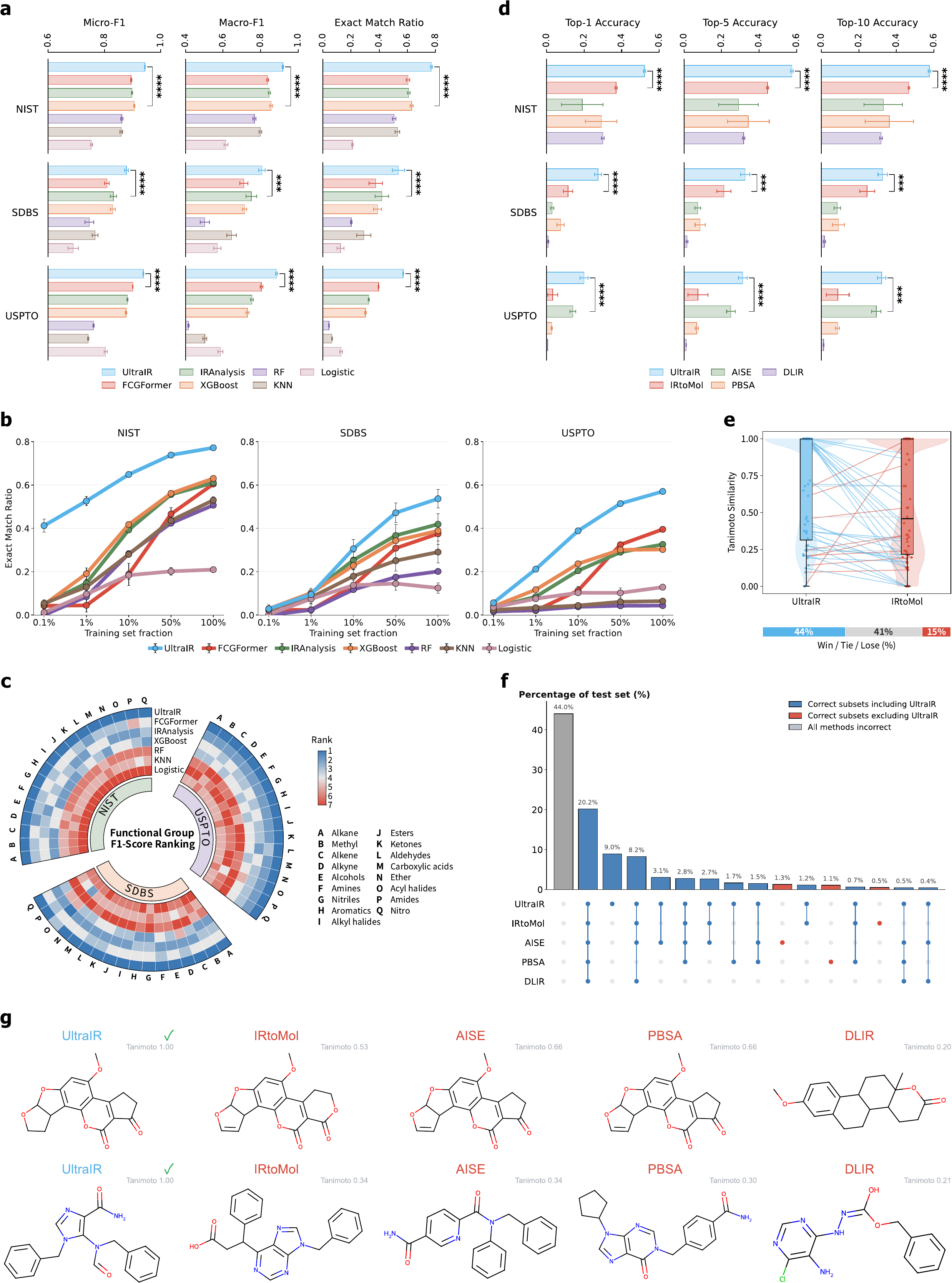}
\end{figure}

We next examined whether UltraIR's performance advantage persisted as the amount of labeled training data varied. Using EMR as the evaluation metric, UltraIR consistently outperformed all competing methods at every training proportion on NIST and USPTO (Fig.~\ref{fig:molecular_interpretation}b). On SDBS, UltraIR performed comparably to the strongest competing method at the two lowest training proportions, 0.1\% and 1\%, and clearly outperformed it at all higher proportions. By contrast, the two task-specific deep-learning models, particularly FCGFormer, underperformed the strongest conventional baseline at the two lowest training proportions and were markedly inferior to UltraIR on NIST and USPTO. These results support the central simulation-to-real premise of UltraIR: large-scale simulated spectra provide transferable chemical information that enables reliable inference across a broad range of labeled-data availability, particularly when downstream supervision is scarce.

Finally, we examined whether the aggregate performance advantage of UltraIR extended across individual functional groups rather than being driven by a small number of frequent labels. Per-group F1-score rankings placed UltraIR consistently among the top-ranked methods across NIST, SDBS, and USPTO (Fig.~\ref{fig:molecular_interpretation}c). This pattern covered 17 functional groups spanning hydrocarbons and unsaturated motifs, oxygen- and nitrogen-containing groups, halogenated groups, and carbonyl-containing derivatives. The aggregate gains therefore reflect broad performance across chemically diverse functional groups rather than gains driven by a narrow subset of common labels. This breadth supports the use of UltraIR for functional interpretation across diverse molecular chemistry.

\subsubsection*{Molecular structure elucidation.}

For molecular structure elucidation, UltraIR was benchmarked against representative IR structure-elucidation models, including IRtoMol, AISE, PBSA, and DLIR \citep{irtomol, ibm, denovo, dlir}. Given an IR spectrum and its molecular formula, each model generates a ranked list of molecular candidates represented as Simplified Molecular Input Line Entry System (SMILES) strings. Performance was evaluated using top-1, top-5, and top-10 accuracy, together with fingerprint-based Tanimoto similarity. The top-\(k\) metrics assess exact structure recovery by determining whether the reference molecule appears among the \(k\) highest-ranked candidates. Fingerprint-based Tanimoto similarity provides a complementary assessment of structural agreement between the top-ranked candidate and the reference molecule, with higher values indicating closer structural similarity. We examined performance from three perspectives: overall exact-recovery accuracy across datasets, structural similarity of top-ranked candidates, and the overlap and characteristics of molecules correctly solved by different methods.

Across NIST, SDBS, and USPTO, UltraIR achieved the strongest overall performance among the compared methods for molecular structure elucidation (Fig.~\ref{fig:molecular_interpretation}d). Its top-1, top-5, and top-10 accuracies show that the learned spectral representation can prioritize the reference structure within a molecular-formula-constrained candidate set. In contrast, the other methods exhibited greater dataset-to-dataset variation in both overall performance and relative ranking.

\begingroup
\captionsetup{type=figure}
\captionof{figure}{
\textbf{UltraIR enables molecular structural interpretation from infrared spectra.}
\textbf{a,} Overall functional-group prediction performance on NIST, SDBS, and USPTO, evaluated using Micro-F1, Macro-F1, and exact match ratio (EMR). EMR counts a prediction as correct only when the complete functional-group set is recovered.
\textbf{b,} Training-data scaling analysis for functional-group prediction, showing EMR as the fraction of labeled training spectra varies from 0.1\% to 100\%.
\textbf{c,} Functional-group-wise F1-score rankings across 17 chemically diverse labels on NIST, SDBS, and USPTO, with lower ranks indicating stronger relative performance for each functional group.
\textbf{d,} Molecular structure elucidation on NIST, SDBS, and USPTO, evaluated using top-1, top-5, and top-10 accuracy. A prediction is counted as correct when at least one of the top \(k\) candidates matches the ground-truth molecular structure after canonicalization.
\textbf{e,} Paired comparison of fingerprint-based Tanimoto similarity between each ground-truth molecule and the corresponding top-1 candidates generated by UltraIR and IRtoMol on the NIST test set. Lines connect predictions for the same molecule.
\textbf{f,} Correct-case overlap among UltraIR, IRtoMol, AISE, PBSA, and DLIR on the NIST test set. The 15 most frequent correct-case subsets are shown together with the fraction of molecules for which all methods were incorrect. Blue and red denote subsets including and excluding UltraIR, respectively, whereas gray denotes cases missed by all methods.
\textbf{g,} Two representative top-1 structure-elucidation examples from the NIST test set, comparing molecular structures predicted by UltraIR and competing methods.
}
\label{fig:molecular_interpretation}
\endgroup

We further assessed the structural similarity of the top-1 candidates on NIST using fingerprint-based Tanimoto similarity. For each spectrum, we compared the Tanimoto similarity between the ground-truth molecule and the top-1 candidates generated by UltraIR and IRtoMol, the strongest competing method on NIST in the exact-recovery evaluation. UltraIR produced candidates with higher, equal, and lower similarity than IRtoMol for 44\%, 41\%, and 15\% of spectra, respectively (Fig.~\ref{fig:molecular_interpretation}e). Thus, UltraIR matched or exceeded IRtoMol in structural similarity for 85\% of the evaluated spectra, and higher-similarity candidates outnumbered lower-similarity candidates by 29 percentage points. Together with the exact-recovery results, this comparison shows that UltraIR not only more often ranks the ground-truth molecule among its leading candidates, but also more often produces a top-ranked candidate that is structurally closer to the ground-truth molecule.

We next analyzed the overlap of correctly solved molecules across UltraIR, IRtoMol, AISE, PBSA, and DLIR on the NIST dataset, focusing on the 15 most frequent correct-case subsets (Fig.~\ref{fig:molecular_interpretation}f). The largest subset comprised molecules that were not solved by any of the five methods, accounting for 44.0\% of the test set. Among the displayed correct-case subsets, those including UltraIR accounted for 52.0\% of test samples, whereas subsets excluding UltraIR accounted for only 2.9\%. Notably, all five methods correctly solved 20.2\% of the test set, while UltraIR alone correctly solved a further 9.0\%. This pattern shows that UltraIR both recovers a substantial set of molecules shared with existing methods and contributes a sizable set of correct predictions not obtained by the competing structure-elucidation approaches.

Representative NIST test cases further illustrate the chemical basis of this advantage (Fig.~\ref{fig:molecular_interpretation}g). In the first case, the ground-truth molecule comprised an oxygen-rich fused polycyclic scaffold containing a methoxy substituent, cyclic ethers, and two ring-associated carbonyl groups. UltraIR recovered the exact top-1 structure with a Tanimoto similarity of 1.00. AISE and PBSA retained most of the fused scaffold but introduced an incorrect double bond within an oxygen-containing ring, each yielding a similarity of 0.66. IRtoMol introduced the same bond-order error and replaced a ring carbon with an oxygen atom, resulting in a similarity of 0.53, whereas DLIR generated a substantially different and less oxygenated polycyclic framework with a similarity of 0.20. In the second case, the ground-truth molecule contained a nitrogen-rich heteroaromatic core bearing carboxamide and formamide functionalities and two benzyl substituents. UltraIR again recovered the exact structure with a similarity of 1.00. By contrast, the competing candidates altered the heteroaromatic core and the connectivity of the carbonyl-containing and aromatic substituents, with DLIR additionally introducing a chlorine atom absent from the ground truth. IRtoMol, AISE, PBSA, and DLIR achieved similarities of only 0.34, 0.34, 0.30, and 0.21, respectively.

Collectively, these results establish UltraIR as a strong approach for functional and structural interpretation of IR spectra, spanning functional-group prediction and molecular structure elucidation.

\subsection*{UltraIR enables physicochemical property prediction from infrared spectra}

Beyond molecular structural interpretation, we evaluated whether UltraIR could predict continuous molecular physicochemical properties from IR spectra. Physicochemical property prediction was evaluated on NIST, SDBS, and USPTO for 11 structure-derived molecular descriptors and scores: synthetic accessibility (SA) score, \(\log P\), topological polar surface area (TPSA), hydrogen-bond donors, hydrogen-bond acceptors, rotatable bonds, the fraction of \(sp^3\)-hybridized carbon atoms (Fraction Csp3), quantitative estimate of drug-likeness (QED), aromatic rings, aliphatic rings, and Bertz complexity (BertzCT). UltraIR was compared with conventional regression baselines, including XGBoost regression, support-vector regression (SVR), KNN regression, and partial least-squares regression (PLSR) \citep{chen2016xgboost, pedregosa2011scikit, wold2001pls}. Performance was evaluated using normalized mean absolute error (normalized MAE), normalized root mean squared error (normalized RMSE), and the coefficient of determination \(R^{2}\). Normalized MAE and normalized RMSE quantify scale-adjusted prediction error, whereas \(R^{2}\) measures the proportion of variance explained by the model. We further analyzed property-wise \(R^{2}\) rankings and BertzCT complexity-thresholded prediction error.

\begin{figure}[!tb]
    \centering
    \includegraphics[width=0.95\textwidth]{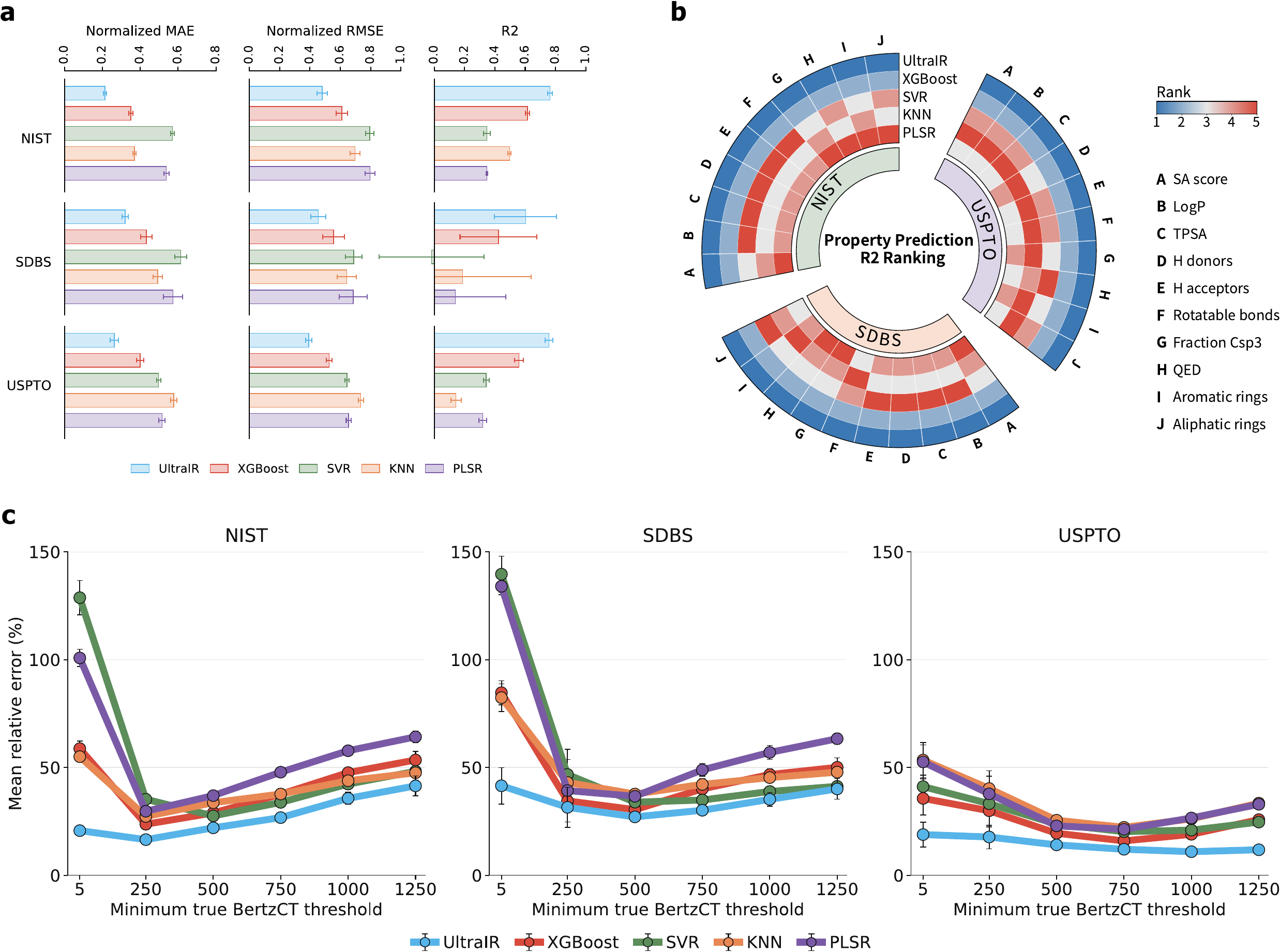}
    \caption{
\textbf{UltraIR enables physicochemical property prediction from infrared spectra.}
\textbf{a,} Overall physicochemical property prediction on NIST, SDBS, and USPTO, evaluated using normalized MAE, normalized RMSE, and \(R^{2}\).
\textbf{b,} Property-wise \(R^{2}\) rankings for 10 of the 11 physicochemical properties across NIST, SDBS, and USPTO. The remaining property, BertzCT, is examined separately in c. Lower ranks indicate stronger relative performance for each property.
\textbf{c,} Mean relative error for BertzCT prediction across cumulative molecular-complexity thresholds on NIST, SDBS, and USPTO. At each threshold, performance is evaluated for molecules with true BertzCT values greater than or equal to that threshold.
    }
    \label{fig:property_prediction}
\end{figure}

Across all three datasets, UltraIR achieved the lowest normalized MAE and normalized RMSE, together with the highest \(R^{2}\) among the evaluated methods (Fig.~\ref{fig:property_prediction}a). XGBoost was the strongest conventional baseline, whereas SVR, KNN, and PLSR showed larger normalized errors or lower \(R^{2}\) values in several settings. Property-wise \(R^{2}\) rankings for the ten properties included in the ranking analysis placed UltraIR first in all 30 property--dataset combinations (Fig.~\ref{fig:property_prediction}b). This consistent ranking shows that UltraIR's advantage extends across the ten ranked properties rather than being driven by a single target.

We next examined BertzCT prediction across cumulative molecular-complexity thresholds. At each threshold, mean relative error was calculated for molecules whose true BertzCT value met or exceeded that threshold. UltraIR achieved the lowest mean relative error at every threshold across NIST, SDBS, and USPTO, whereas conventional regression baselines showed larger errors and more pronounced threshold-dependent variation, particularly for high-complexity subsets (Fig.~\ref{fig:property_prediction}c). These results indicate that UltraIR maintains accurate continuous-property prediction across a broad range of molecular complexity.

Together, these results extend the molecular-level interpretation capability of UltraIR beyond discrete functional-group labels and molecular structure elucidation. The learned IR representations capture continuous physicochemical attributes across diverse molecular properties and complexity regimes.

\subsection*{UltraIR enables mixture-component identification and quantification from infrared spectra}

We evaluated targeted component detection and fractional contribution estimation across NIST, SDBS, and USPTO. Targeted component detection under increasing mixture complexity was further evaluated using the DeepMIR liquid-mixture dataset \citep{deepmir}, whereas mixture-level component quantification was assessed using FTIR mixture datasets \citep{digital_discovery_2022}.

\subsubsection*{Targeted component detection and fractional contribution estimation.}

For pairwise mixture analysis, we evaluated UltraIR on targeted component detection and targeted fractional contribution estimation across NIST, SDBS, and USPTO. In both tasks, the input consisted of a reference single-compound spectrum and a mixture spectrum. For targeted component detection, UltraIR was benchmarked against DeepMIR \citep{deepmir}, reverse match, and the hit quality index (HQI). Performance was evaluated using accuracy, Macro-F1, and the area under the receiver operating characteristic curve (ROC-AUC). For targeted fractional contribution estimation, UltraIR was compared with a DeepMIR-derived regression baseline and conventional regression baselines, including XGBoost regression, support-vector regression (SVR), KNN regression, and partial least-squares regression (PLSR). Performance was evaluated using mean absolute error (MAE), root mean squared error (RMSE), and \(R^{2}\).

\begin{figure}[!t]
    \centering
    \includegraphics[width=0.95\textwidth]{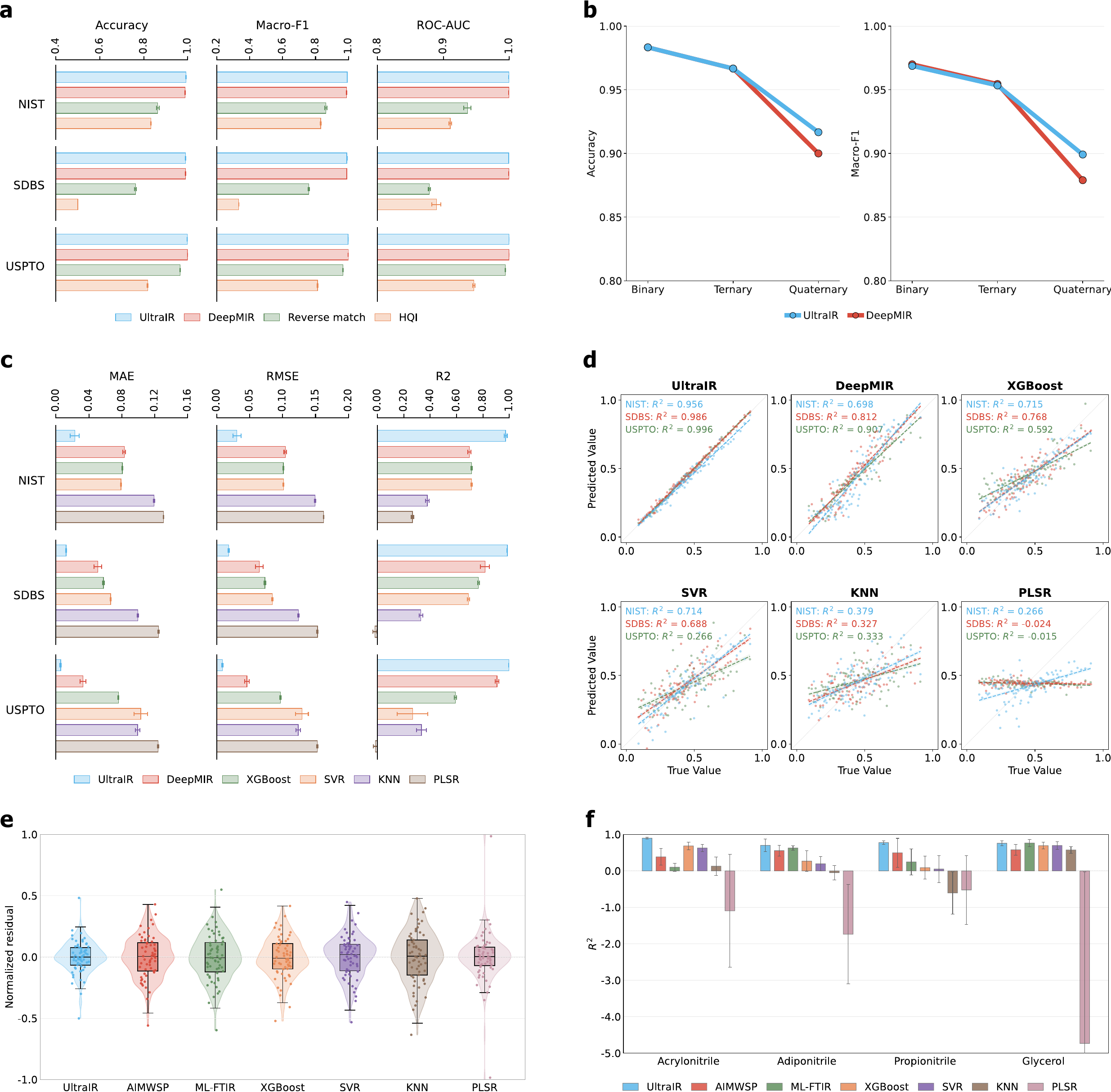}
    \caption{
    \textbf{UltraIR enables mixture-component identification and quantification from infrared spectra.}
\textbf{a,} Targeted component detection on NIST, SDBS, and USPTO, evaluated using accuracy, Macro-F1, and ROC-AUC. UltraIR is compared with DeepMIR, reverse match, and hit quality index (HQI).
\textbf{b,} Targeted component detection in binary, ternary, and quaternary mixtures, comparing UltraIR and DeepMIR using accuracy and Macro-F1.
\textbf{c,} Targeted fractional contribution estimation on NIST, SDBS, and USPTO, evaluated using mean absolute error (MAE), root mean squared error (RMSE), and \(R^{2}\). 
\textbf{d,} Parity plots for targeted fractional contribution estimation, comparing predicted and true target-component contributions for UltraIR and competing methods across NIST, SDBS, and USPTO. The identity line denotes ideal agreement, and dashed lines show fitted relationships.
\textbf{e,} Normalized residual distributions for mixture-level multi-component quantification from mixture spectra.
\textbf{f,} Component-wise \(R^{2}\) values for mixture-level quantification of acrylonitrile, adiponitrile, propionitrile, and glycerol.
    }
    \label{fig:mixture_analysis}
\end{figure}

UltraIR achieved strong targeted component-detection performance across all three datasets (Fig.~\ref{fig:mixture_analysis}a). On NIST, UltraIR attained higher accuracy and Macro-F1 than DeepMIR while achieving comparable ROC-AUC. On SDBS and USPTO, UltraIR and DeepMIR performed comparably across all three metrics. Across all datasets, UltraIR consistently outperformed the direct spectral-matching baselines: reverse match and HQI.

We next examined whether this performance persisted as mixture complexity increased. UltraIR and DeepMIR performed comparably for binary and ternary mixtures, whereas UltraIR established a clearer advantage in both accuracy and Macro-F1 for quaternary mixtures (Fig.~\ref{fig:mixture_analysis}b). This pattern indicates that UltraIR retains stronger target-component detection as additional components increase spectral overlap and make the target contribution more difficult to isolate.

UltraIR also achieved the strongest performance in targeted fractional contribution estimation across all three datasets, substantially and significantly outperforming the strong DeepMIR-derived regression baseline. It attained the lowest MAE and RMSE, together with the highest \(R^{2}\) among all evaluated methods (Fig.~\ref{fig:mixture_analysis}c). Its \(R^{2}\) values reached 0.956 on NIST, 0.986 on SDBS, and 0.996 on USPTO (Fig.~\ref{fig:mixture_analysis}d). The parity plots showed that UltraIR predictions closely followed the identity line, whereas competing methods exhibited larger deviations and weaker calibration (Fig.~\ref{fig:mixture_analysis}d). These results establish that UltraIR supports both qualitative identification of a target component and quantitative estimation of its contribution to a mixture spectrum.

\subsubsection*{Mixture-level component quantification.}

We finally evaluated mixture-level component quantification, in which the model receives only a mixture spectrum and predicts the proportions of multiple components. This task followed a simulation-to-experiment transfer protocol: models were first trained on a simulated FTIR mixture dataset and then fine-tuned and evaluated only on experimentally acquired FTIR mixture spectra \citep{digital_discovery_2022}. UltraIR was compared with the mixture-analysis models AIMWSP and ML-FTIR \citep{acsnano_2023_inspired, digital_discovery_2022}, as well as conventional regression baselines including XGBoost regression, SVR, KNN regression, and PLSR. Performance was assessed using normalized residual distributions and component-wise \(R^{2}\).

UltraIR produced residuals most tightly concentrated around zero, indicating the most accurate and stable mixture-level concentration estimates among the evaluated methods (Fig.~\ref{fig:mixture_analysis}e). Component-wise analysis further confirmed this advantage. UltraIR achieved the highest \(R^{2}\) for each of the four quantified components: acrylonitrile, adiponitrile, propionitrile, and glycerol (Fig.~\ref{fig:mixture_analysis}f). Several competing methods showed pronounced component-dependent degradation. PLSR yielded negative \(R^{2}\) values across all four components, and KNN also produced a negative \(R^{2}\) for propionitrile, indicating performance below that of a mean-value predictor in these settings.

Together, these results show that UltraIR extends molecular-level IR analysis from individual compounds to chemical mixtures. It enables reference-guided component detection and fractional contribution estimation, as well as multi-component quantification from a single mixture spectrum. UltraIR thus supports functional-group prediction, molecular structure elucidation, physicochemical property prediction, and mixture analysis.

\subsection*{UltraIR supports biological and botanical infrared sensing}

Having established UltraIR across molecular-level tasks, we next examined whether it could transfer to biological and botanical IR sensing tasks involving experimentally acquired spectra from complex real-world samples. We considered two applications with distinct sample types and analytical objectives: genus-level bacterial classification and medicinal-herb characterization, encompassing geographic origin traceability and chemical constituent quantification.

\subsubsection*{Bacterial classification.}

For bacterial classification (Fig.~\ref{fig:biological_botanical_sensing}a), UltraIR was evaluated on FTIR spectra of fast-growing bacterial isolates from green snow \citep{bacteria}. The task was formulated as genus-level classification and benchmarked against conventional machine-learning baselines, including XGBoost, RF, KNN, and a logistic classifier. Performance was evaluated using accuracy, Macro-F1, and the Matthews correlation coefficient (MCC), which provides a robust measure under class imbalance.

UltraIR achieved the highest accuracy, Macro-F1, and MCC among all compared methods (Fig.~\ref{fig:biological_botanical_sensing}c). Genus-wise F1 analysis further showed that UltraIR consistently outperformed the conventional baselines across all nine evaluated genera, whereas baseline performance varied substantially among genera (Fig.~\ref{fig:biological_botanical_sensing}d). Thus, UltraIR's aggregate advantage reflected broad genus-level discrimination rather than performance gains confined to a small subset of genera.

\begin{figure}[!htbp]
    \centering
    \includegraphics[width=0.94\textwidth]{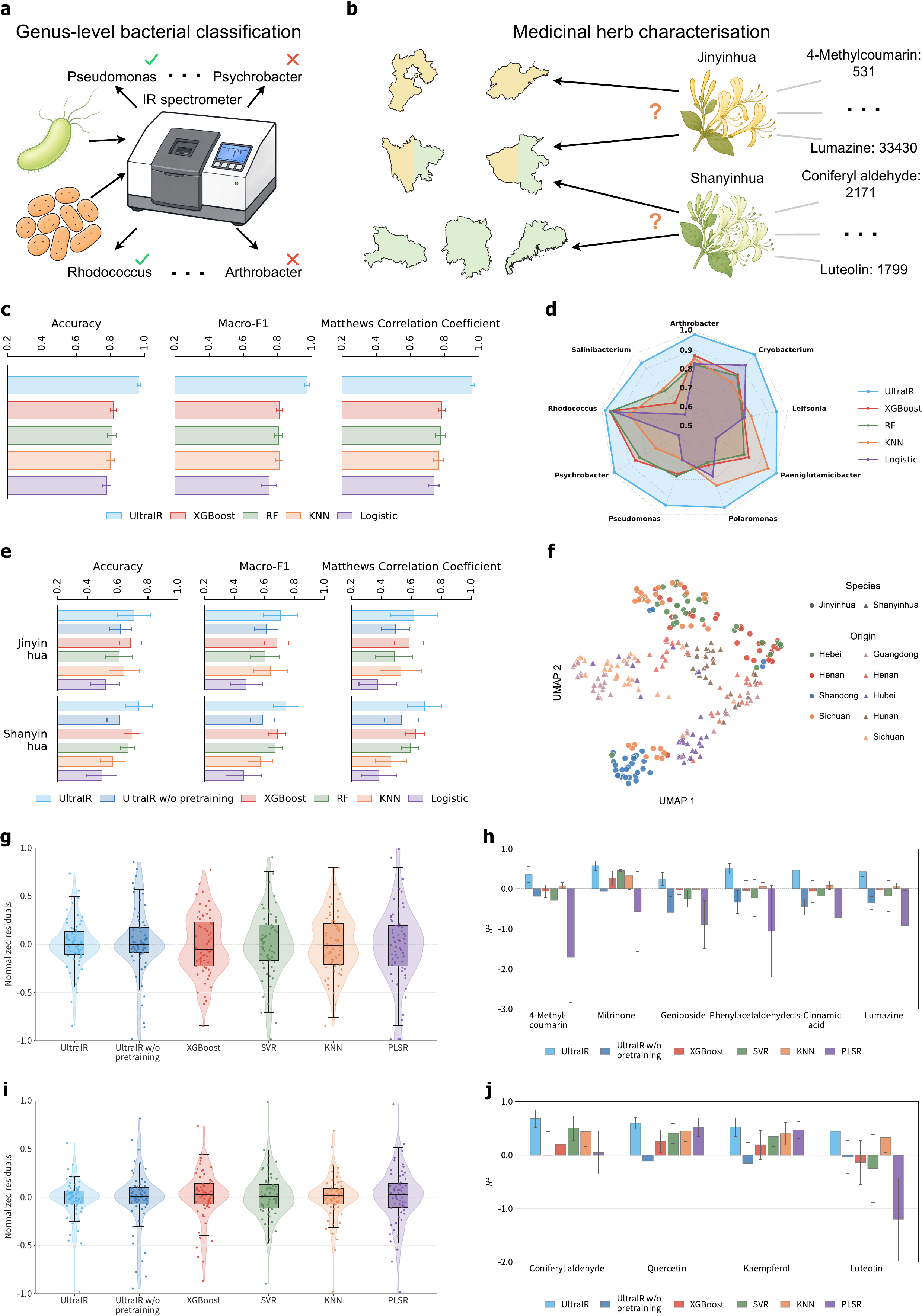}
\end{figure}

\subsubsection*{Medicinal herb characterization.}

We next evaluated UltraIR on two newly generated in-house experimental datasets of Jinyinhua (\textit{Lonicerae Japonicae Flos}) and Shanyinhua (\textit{Lonicerae Flos}), comprising newly acquired IR spectra for both geographic origin traceability and chemical constituent quantification (Fig.~\ref{fig:biological_botanical_sensing}b). The origin-traceability datasets comprised 120 Jinyinhua spectra from Shandong, Henan, Hebei, and Sichuan and 150 Shanyinhua spectra from Hunan, Hubei, Sichuan, Henan, and Guangdong, with 30 spectra per origin. Liquid chromatography--mass spectrometry (LC-MS)-derived relative abundances of the target constituents, expressed in arbitrary units (a.u.), served as regression labels for 60 Jinyinhua and 75 Shanyinhua samples. Details of the in-house datasets, including dataset composition, LC-MS sample preparation, and data acquisition, are provided in the Appendix. We retained a matched UltraIR model without simulated pretraining as an ablation for these tasks. For geographic origin traceability, UltraIR and the no-pretraining ablation were compared with XGBoost, RF, KNN, and a logistic classifier using accuracy, Macro-F1, and MCC. For chemical constituent quantification, they were compared with XGBoost regression, SVR, KNN regression, and PLSR using normalized residual distributions and constituent-wise \(R^{2}\).

For geographic origin traceability, UltraIR achieved the highest mean accuracy, Macro-F1 and MCC on both Jinyinhua and Shanyinhua, outperforming all conventional baselines (Fig.~\ref{fig:biological_botanical_sensing}e). By contrast, the no-pretraining ablation underperformed UltraIR across all three metrics on both Jinyinhua and Shanyinhua and also fell below XGBoost, the strongest conventional baseline, in every comparison. A Uniform Manifold Approximation and Projection (UMAP) of embeddings extracted by UltraIR provided a complementary qualitative view of the learned spectral organization (Fig.~\ref{fig:biological_botanical_sensing}f). Spectra from the same geographic origin largely co-localized within each herb species, while Jinyinhua and Shanyinhua occupied largely distinct regions of the embedding space rather than being broadly intermixed. This organization emerged even though herb species identity was not used as a supervised label in the origin-traceability task and is consistent with UltraIR encoding spectral variation associated with both medicinal-herb identity and geographic origin.

We next moved from categorical origin assignment to quantitative estimation of chemical constituents. For both Jinyinhua and Shanyinhua, UltraIR produced normalized residual distributions that were more tightly concentrated around zero than those of the no-pretraining ablation and the conventional regression baselines, with fewer large residuals (Fig.~\ref{fig:biological_botanical_sensing}g,i). The no-pretraining ablation nevertheless produced a compact interquartile residual distribution. Its interquartile range was narrower than that of KNN regression, the strongest conventional baseline by this criterion, on Jinyinhua and comparable to that of KNN regression on Shanyinhua. However, the ablation also produced more large-residual outliers.

Because residual centering and dispersion do not establish how much constituent-specific variation a model explains, we next examined constituent-wise \(R^{2}\). UltraIR achieved the highest \(R^{2}\) for every evaluated constituent in both Jinyinhua and Shanyinhua, with particularly large gains over the no-pretraining ablation (Fig.~\ref{fig:biological_botanical_sensing}h,j). Several conventional regression baselines showed marked constituent-dependent degradation and, in some cases, negative \(R^{2}\) values. The no-pretraining ablation also showed weak constituent-level generalization despite its residual distributions being centered close to zero. On Jinyinhua, its \(R^{2}\) values were negative for all six constituents and exceeded only those of PLSR, the weakest method on this dataset. On Shanyinhua, it produced the lowest \(R^{2}\) among all methods for three of the four constituents. The consistent contrast between UltraIR and the matched ablation provides direct evidence that simulated pretraining was critical under limited labeled supervision.

\begingroup
\captionsetup{type=figure}
\captionof{figure}{
\textbf{UltraIR supports biological and botanical infrared sensing.}
\textbf{a,} Schematic of genus-level bacterial classification from infrared spectra.
\textbf{b,} Schematic of medicinal-herb characterization using in-house experimental datasets, comprising geographic origin traceability and chemical constituent quantification for Jinyinhua and Shanyinhua.
\textbf{c,} Overall genus-level bacterial classification, evaluated using accuracy, Macro-F1, and the Matthews correlation coefficient (MCC).
\textbf{d,} Genus-wise F1 scores across the nine evaluated bacterial genera.
\textbf{e,} Geographic origin traceability of Jinyinhua and Shanyinhua, evaluated using accuracy, Macro-F1, and MCC. UltraIR is compared with its matched no-pretraining ablation and conventional classifiers.
\textbf{f,} Uniform Manifold Approximation and Projection (UMAP) of medicinal-herb spectral embeddings extracted by UltraIR. Points are colored by geographic origin, and marker shapes indicate herb species.
\textbf{g,} Normalized residual distributions for chemical constituent quantification in Jinyinhua, comparing UltraIR, its matched no-pretraining ablation, and conventional regression baselines.
\textbf{h,} Constituent-wise \(R^{2}\) values for the six quantified Jinyinhua constituents across the same methods.
\textbf{i,} Normalized residual distributions for chemical constituent quantification in Shanyinhua, comparing UltraIR, its matched no-pretraining ablation, and conventional regression baselines.
\textbf{j,} Constituent-wise \(R^{2}\) values for the four quantified Shanyinhua constituents across the same methods.
}
\label{fig:biological_botanical_sensing}
\endgroup

Together with the bacterial classification results, these findings show that UltraIR supports IR analysis of complex biological and botanical samples across genus-level identification, medicinal-herb geographic origin traceability, and chemical constituent quantification.

\subsection*{UltraIR supports environmental infrared sensing}

Having established UltraIR's performance in biological and botanical IR sensing, we finally evaluated the model for environmental IR sensing using experimentally acquired spectra from complex samples. We considered two representative environmental applications: microplastics classification and soil physicochemical property prediction.

\subsubsection*{Microplastics classification.}

For microplastics classification (Fig.~\ref{fig:environmental_sensing}a), UltraIR was evaluated on IR spectra of environmentally sourced microplastics \citep{microp}. UltraIR was compared with task-specific deep-learning baselines, including Softmax and DB-CNN-CBAM \citep{microp, dbcnn}, and conventional machine-learning baselines, including XGBoost, RF, KNN, and a logistic classifier. Performance was evaluated using accuracy, Macro-F1, and MCC. We further analyzed the true-class margin, defined as the correct-class logit minus the highest logit among the non-true classes. Larger positive values indicate stronger separation of the correct class from its nearest competitor.

Across accuracy, Macro-F1, and MCC, UltraIR achieved the strongest performance among all evaluated methods (Fig.~\ref{fig:environmental_sensing}b). Its advantage over both task-specific deep-learning and conventional machine-learning baselines demonstrates effective transfer from molecular benchmarks to polymer identification in complex environmental spectra.

\begin{figure}[p] 
    \centering       
    \includegraphics[width=0.94\textwidth]{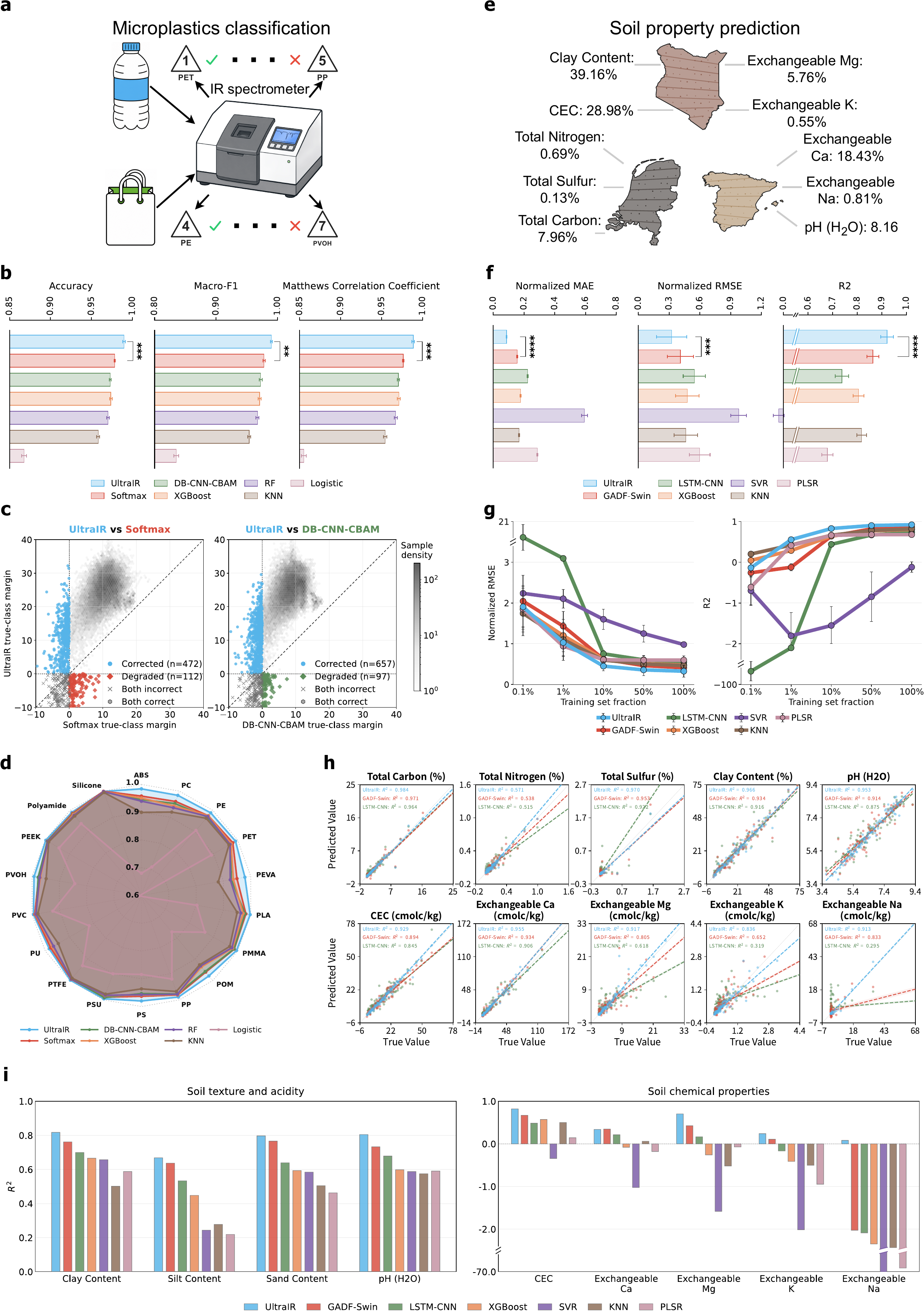} 
\end{figure}

We next examined whether this aggregate performance advantage was accompanied by larger true-class margins. In the pairwise margin plots, many spectra lay above the diagonal, indicating that UltraIR assigned a larger true-class margin than Softmax or DB-CNN-CBAM, including numerous cases in which both methods correctly predicted the polymer class (Fig.~\ref{fig:environmental_sensing}c). UltraIR also corrected substantially more baseline errors than it introduced. Relative to Softmax, UltraIR corrected 472 baseline errors while introducing 112 errors. Relative to DB-CNN-CBAM, it corrected 657 baseline errors while introducing 97 errors. Thus, UltraIR not only corrected more misclassified spectra, but also achieved stronger separation between the correct polymer class and its nearest alternatives.

Finally, class-wise F1 analysis showed that UltraIR maintained consistently high performance across all 18 polymer categories, including common commodity polymers and more specialized materials (Fig.~\ref{fig:environmental_sensing}d). This broad class-wise advantage shows that the overall improvement was not driven by a narrow subset of readily distinguishable polymers, but extended across the microplastics classification label space.

\subsubsection*{Soil property prediction.}

For soil property prediction (Fig.~\ref{fig:environmental_sensing}e), we evaluated whether UltraIR could estimate continuous physicochemical properties from soil IR spectra using the Open Soil Spectral Library (OSSL) \citep{ossl}. UltraIR was compared with soil spectral-modeling baselines, including GADF-Swin and LSTM-CNN \citep{gadfswin, lstmcnn}, as well as conventional regression baselines, including XGBoost regression, SVR, KNN regression, and PLSR. Performance was evaluated using normalized MAE and RMSE, together with property-wise \(R^{2}\), across ten soil properties: total carbon, total nitrogen, total sulfur, clay content, pH \((\mathrm{H_2O})\), cation-exchange capacity (CEC), and exchangeable Ca, Mg, K, and Na.

\begingroup
\captionsetup{type=figure}
\captionof{figure}{
\textbf{UltraIR supports environmental infrared sensing across microplastics and soil analysis.}
\textbf{a,} Schematic of microplastics classification from infrared spectra of environmentally sourced microplastics.
\textbf{b,} Overall microplastics classification, evaluated using accuracy, Macro-F1, and the Matthews correlation coefficient (MCC).
\textbf{c,} Pairwise true-class margin comparisons for UltraIR versus Softmax and UltraIR versus DB-CNN-CBAM. The true-class margin is defined as the correct-class logit minus the highest logit among the non-true classes, with points above the diagonal indicating a larger margin for UltraIR. Corrected and degraded samples denote cases classified correctly by UltraIR but incorrectly by the comparator, and vice versa, respectively.
\textbf{d,} Class-wise F1 scores across 18 polymer categories.
\textbf{e,} Schematic of soil physicochemical property prediction from infrared spectra.
\textbf{f,} Overall soil physicochemical property prediction, evaluated using normalized mean absolute error (normalized MAE), normalized root mean squared error (normalized RMSE), and \(R^{2}\).
\textbf{g,} Training-data scaling analysis for soil physicochemical property prediction, reporting normalized RMSE and \(R^{2}\) as the labeled training fraction varies from 0.1\% to 100\%.
\textbf{h,} Parity plots for soil property prediction, comparing predicted and measured values for UltraIR, GADF-Swin, and LSTM-CNN across ten properties: total carbon, total nitrogen, total sulfur, clay content, pH \((\mathrm{H_2O})\), cation-exchange capacity (CEC), and exchangeable Ca, Mg, K, and Na. The identity line denotes ideal agreement.
\textbf{i,} Cross-instrument and cross-laboratory soil property prediction. Bars show property-wise \(R^{2}\) for soil texture and acidity and for soil chemical properties.
}
\label{fig:environmental_sensing}
\endgroup

Across the three aggregate regression metrics, UltraIR achieved the strongest overall performance in soil property prediction (Fig.~\ref{fig:environmental_sensing}f). It attained the lowest normalized MAE and normalized RMSE and the highest \(R^{2}\) among all evaluated methods.

We further assessed performance under reduced labeled-data settings. As the training-set fraction decreased from 100\% to 0.1\%, normalized RMSE increased and \(R^{2}\) decreased across all methods (Fig.~\ref{fig:environmental_sensing}g). At the 1\% training fraction, UltraIR achieved lower normalized RMSE than XGBoost regression and KNN regression and attained the highest \(R^{2}\) among all evaluated methods, although its normalized RMSE was slightly higher than that of PLSR. At this fraction, UltraIR substantially outperformed the task-specific neural baselines GADF-Swin and LSTM-CNN, with LSTM-CNN showing particularly severe performance degradation. As the labeled training fraction increased to 10\%, 50\%, and 100\%, both task-specific models achieved progressively lower normalized RMSE and higher \(R^{2}\), but remained inferior to UltraIR. This strong performance with only 1\% of the labeled training data demonstrates the data efficiency of UltraIR for soil spectroscopy, which is practically important because soil reference measurements are expensive to acquire and are often unevenly available across soil properties and sampling regions.

Property-wise analysis further confirmed the consistency of this advantage. UltraIR achieved the highest \(R^{2}\) for all ten soil properties (Fig.~\ref{fig:environmental_sensing}h). Its predictions closely followed the identity line for total carbon, total sulfur, clay content, exchangeable calcium, and pH, for which it achieved particularly strong performance. Total nitrogen remained the most challenging property across the evaluated methods, yet UltraIR still achieved the highest \(R^{2}\). These results demonstrate broad predictive capability across soil attributes with distinct chemical origins, concentration ranges, and spectral signatures.

Together with the microplastics results, these findings show that UltraIR supports environmental IR sensing across both discrete material classification and continuous quantitative estimation. Its accuracy, robustness under limited labeled data, and consistent property-wise performance support reliable IR-based analysis of complex environmental samples.

\subsubsection*{Soil property prediction as a case study of cross-instrument and cross-laboratory generalization}

To examine UltraIR's capacity to generalize across FTIR instruments and laboratories, we used soil property prediction as a case study. UltraIR was compared with the same soil spectral-modeling baselines: GADF-Swin, LSTM-CNN, XGBoost regression, SVR, KNN regression, and PLSR. Relative to the preceding experiment, this comparison used different subsets of soil IR spectra from OSSL \citep{ossl} and a partially different set of soil properties. Performance was evaluated using property-wise \(R^{2}\) for clay content, silt content, sand content, pH \((\mathrm{H_2O})\), CEC, and exchangeable Ca, Mg, K, and Na.

All models were trained using spectra from the Kellogg Soil Survey Laboratory (KSSL) subset of OSSL and evaluated using zero-shot target-domain inference on the ICRAF--ISRIC subset of OSSL. The source and target subsets differed in laboratory, FTIR instrument, and sample preparation procedures. Detailed metadata and label distributions for both subsets are provided in Appendix Tables~\ref{tab:crosslab_soil_dataset_metadata} and \ref{tab:crosslab_soil_label_distribution}.

UltraIR achieved the highest \(R^{2}\) for all soil texture and acidity properties and for most soil chemical properties (Fig.~\ref{fig:environmental_sensing}i). UltraIR outperformed all competing methods for CEC and exchangeable Mg, K, and Na, while performing comparably to GADF-Swin and better than the remaining methods for exchangeable Ca. UltraIR was also the only method to maintain positive \(R^{2}\) across all evaluated properties. Performance differences were most pronounced for the chemical properties, for which several baselines yielded negative \(R^{2}\), particularly for exchangeable Na. This case study demonstrates that UltraIR achieved more stable and generally stronger performance during zero-shot inference across instruments and laboratories, which represents an important practical challenge in real-world chemical sensing.

\section{Discussion}
\label{sec:discussion}

IR spectroscopy is widely used for chemical sensing, but extracting reliable chemical information from complex spectra at scale remains difficult across analytical objectives and datasets. UltraIR addresses this challenge by enabling simulation-to-real transfer learning through a foundation model for IR spectroscopy with more than 100 million parameters. The model learns a shared spectral representation by pretraining on approximately 60 million simulated IR spectra using complementary objectives for spectral reconstruction, molecular fingerprint similarity alignment, and functional-group prediction. The pretrained encoder is then adapted to each downstream analytical objective using task-specific supervision. This design separates large-scale spectral representation learning from downstream adaptation, allowing the same pretrained encoder to support diverse forms of IR chemical sensing and analysis.

A central advance of UltraIR lies in the breadth of chemical inference supported by its shared pretrained encoder. The evaluated tasks span molecular structural interpretation, physicochemical property prediction, mixture-component identification and quantification, and real-world chemical sensing in biological and botanical samples and complex environmental samples, with two newly generated in-house experimental datasets used for botanical sensing. These tasks differ in their inputs, output spaces, and analytical objectives, encompassing multi-label classification, molecular generation, regression, reference-guided mixture analysis, and multi-component quantification. Across these evaluations, UltraIR outperformed conventional machine-learning and task-specific deep-learning baselines. This consistent performance across tasks and datasets indicates that the learned representation can transfer after supervised adaptation rather than remaining tied to a single label space or dataset. UltraIR therefore provides a practical basis for reusing a shared pretrained encoder with a task-specific output module for each analytical objective.

The practical value of UltraIR's simulation-to-real transfer learning was particularly evident in two challenging real-world chemical sensing scenarios: downstream adaptation with limited labeled experimental spectra and zero-shot target-domain inference for the same analytical task across FTIR instruments and laboratories. In training-data scaling experiments, UltraIR retained stronger performance than task-specific neural baselines as labeled supervision decreased. The medicinal-herb ablation further showed that simulated pretraining benefited complex sample analysis under limited labeled supervision. A complementary soil property prediction case study showed that UltraIR maintained more stable and generally stronger performance than competing methods during zero-shot target-domain inference across instruments and laboratories. Simulated pretraining provides broad chemical exposure, whereas task-specific labels align the shared representation with individual analytical objectives, together supporting data-efficient learning and generalization across measurement conditions.

Although UltraIR demonstrated strong performance across diverse infrared chemical analysis tasks, several limitations remain and point to directions for further improvement. UltraIR still requires supervised adaptation and a task-specific output module for each analytical objective. The present study therefore demonstrates broad transferability across analytical objectives after adaptation, rather than zero-shot or universal inference across tasks. A simulation-to-real domain gap also remains because simulated IR spectra cannot fully reproduce experimental variation. This variation can arise from instrument response, spectral resolution, measurement geometry, sample state, preparation procedures, temperature, concentration, scattering, complex matrices, and baseline artifacts. Improving simulation fidelity will require explicit modeling of these factors. Future adaptation strategies could exploit unlabeled experimental IR spectra to reduce the need for labeled experimental spectra.

Overall, UltraIR establishes a scalable framework for simulation-to-real transfer learning in foundation modeling for IR spectroscopy. By coupling pretraining on chemically diverse simulated spectra with supervised downstream adaptation, UltraIR supports chemical sensing and analysis across analytical objectives and datasets, spanning molecules and mixtures, biological and botanical samples, and complex environmental samples. This framework provides a route beyond collections of independently trained task-specific models toward adaptable and data-efficient IR-based chemical sensing systems built on reusable spectral representations.
\newpage
\section{Methods}
\label{sec:method}

\subsection*{Overview of UltraIR}

UltraIR is a foundation model for IR spectroscopy with more than 100 million parameters that enables simulation-to-real transfer learning for chemical sensing and analysis from molecules to complex samples. The framework comprises a shared spectral encoder and task-specific output modules. During pretraining, the encoder is optimized on approximately 60 million simulated IR spectra to learn a shared, chemically informative spectral representation. For downstream adaptation, the pretrained encoder is paired with a newly initialized task-specific output module, and both components are jointly optimized using labeled spectra for the corresponding task. The encoder architecture and pretrained parameters provide a common initialization across downstream objectives, whereas the output module, output dimensionality, and supervised training objective are defined separately for each task.

In UltraIR, the shared encoder is initialized from the pretrained checkpoint. In designated ablation analyses, we additionally evaluated a matched no-pretraining control in which the same encoder architecture and task-specific output modules were initialized randomly and trained using the same downstream data and optimization procedure. Formally, given a preprocessed IR spectrum \(\mathbf{x}\), the shared encoder produces a latent representation

\begin{equation}
\mathbf{z}=f_{\theta}(\mathbf{x}),
\end{equation}

where \(f_{\theta}\) denotes the shared encoder with parameters \(\theta\). For UltraIR, \(\theta\) is initialized from the pretrained parameters \(\theta_{\mathrm{pre}}\). For the no-pretraining control, \(\theta\) is instead initialized randomly. A task-specific output module \(h_{\phi}^{(t)}\) then transforms the shared representation, together with any auxiliary input required by task \(t\), into the corresponding analytical output:

\begin{equation}
\hat{\mathbf{y}}^{(t)}
=
h_{\phi}^{(t)}
\left(
\mathbf{z},\mathbf{u}^{(t)}
\right),
\end{equation}

where \(\mathbf{u}^{(t)}\) denotes optional task-specific inputs, such as a molecular formula or a reference single-compound spectrum. The downstream outputs include functional-group annotations, formula-conditioned molecular structure candidates, molecular physicochemical property estimates, mixture-component detection and quantification outputs, bacterial genus labels, medicinal-herb geographic origin labels and constituent relative abundances, polymer-type labels for microplastics, and soil physicochemical property estimates.

\subsection*{Spectral preprocessing and data augmentation}

Before being supplied to the model, all IR spectra were converted to a common spectral range, sampling grid, and intensity scale. The same deterministic preprocessing pipeline was used for large-scale pretraining on simulated spectra and downstream adaptation on experimental spectra. Stochastic augmentation was stage-specific. A broader set of augmentation operations was used during pretraining, whereas only weak, validation-selected augmentations were considered during downstream adaptation.

\subsubsection*{Standard spectral preprocessing pipeline.}

Each spectrum was first cropped to the mid-IR range of \SIrange{400}{4000}{\per\centi\metre}. Spectra that did not cover the complete range were zero-padded at the missing boundaries. The resulting spectrum was resampled by piecewise-linear interpolation onto a canonical 3,600-point grid and then min--max normalized to the interval \([0, 1]\). This canonical representation was subsequently mapped to the fixed input length required by each model implementation.

\subsubsection*{Pretraining augmentation.}

During pretraining, four stochastic augmentation operations were applied to each spectrum. Additive Gaussian noise was sampled at a random signal-to-noise ratio to represent electronic measurement noise. A random shift along the wavenumber axis represented calibration uncertainty. An additive intensity offset represented baseline drift. Finally, a random subset of spectral positions was masked by setting the corresponding intensities to zero. This masking objective encouraged the encoder to use information from the broader spectral context rather than relying exclusively on isolated local features.

\subsubsection*{Downstream adaptation augmentation.}

For downstream adaptation, stochastic augmentation was not imposed as a universal part of the training pipeline. Instead, weak augmentations, including additive noise, small wavenumber shifts, and baseline offsets, were enabled only when selected using the validation split for the corresponding model--task setting. Random spectral masking was not used during downstream adaptation because most downstream objectives require access to the complete experimental spectral profile.

\subsection*{Hybrid convolutional--Transformer architecture}

As illustrated in Fig.~\ref{fig:task_landscape}b, UltraIR uses a shared hybrid convolutional--Transformer encoder to map an IR spectrum to a latent spectral representation \(\mathbf{z}\) for downstream chemical analysis. The encoder comprises three principal modules: a derivative-aware multi-channel input module, a hierarchical convolutional backbone with multi-scale feature fusion, and a patch-based Transformer encoder. Together, these modules combine derivative-aware spectral feature extraction, multi-scale local feature aggregation, and long-range spectral-context modeling.

\subsubsection*{Derivative-aware multi-channel input module.}

Given the preprocessed IR spectrum $\mathbf{x} \in \mathbb{R}^{1 \times L}$, where $L$ denotes the number of spectral data points, this module augments the original absorbance signal with derivative-based spectral representations to enhance sensitivity to spectral line shape features. Specifically, $\mathbf{x}$ is first smoothed via a Savitzky--Golay (SG) filter to yield $\tilde{\mathbf{x}}$. Two learnable derivative convolutions then compute the first- and second-order spectral derivatives $\mathbf{d}_1$ and $\mathbf{d}_2$, where the corresponding kernels $\mathbf{w}_1$ and $\mathbf{w}_2$ are initialized as first- and second-order central-difference operators, respectively, and remain trainable throughout optimization. Each derivative signal is subsequently normalized using layer normalization, and the three signals are concatenated to form the three-channel representation $\mathbf{x}_{\text{cat}} = [\mathbf{x},\,\mathbf{d}_1,\,\mathbf{d}_2] \in \mathbb{R}^{3 \times L}$. To adaptively reweight each spectral channel, $\mathbf{x}_{\text{cat}}$ is passed through a gated input fusion module that computes a position-wise, input-dependent channel gate:

\begin{equation}
    \mathbf{g} = \sigma\!\left(\mathbf{W}_2^{\text{g}}\,\text{GELU}\!\left(\mathbf{W}_1^{\text{g}}\,\mathbf{x}_{\text{cat}}\right)\right) \in \mathbb{R}^{3 \times L},
\end{equation}

where $\mathbf{W}_1^{\text{g}}$ and $\mathbf{W}_2^{\text{g}}$ are $1{\times}1$ convolutional projections, GELU denotes the Gaussian error linear unit activation, and $\sigma(\cdot)$ denotes the sigmoid function. The fused multi-channel representation is then obtained via element-wise modulation:

\begin{equation}
    \mathbf{x}_{\text{multi}} = \mathbf{x}_{\text{cat}} \odot \mathbf{g} \in \mathbb{R}^{3 \times L},
\end{equation}

where $\odot$ denotes element-wise multiplication.

\subsubsection*{Convolutional spectral encoder.}
The fused representation $\mathbf{x}_{\text{multi}} \in \mathbb{R}^{3 \times L}$ is fed into a convolutional encoder that hierarchically extracts local spectral features across multiple spectral scales. Prior to feature extraction, $\mathbf{x}_{\text{multi}}$ is rescaled by a learnable per-channel scale vector $\boldsymbol{\gamma} \in \mathbb{R}^{3}$, broadcast along the spectral dimension and initialized as $[1.0,\,0.5,\,0.5]^{\top}$ to encode the inductive bias that the original absorbance signal should initially dominate over the two derivative channels, while allowing the network to adjust this balance during training, yielding $\hat{\mathbf{x}}_{\text{multi}} \in \mathbb{R}^{3 \times L}$. The encoder then begins with a convolutional stem that applies a wide-kernel convolution followed by batch normalization and a GELU activation to $\hat{\mathbf{x}}_{\text{multi}}$, mapping it to an initial feature map $\mathbf{H}_0 \in \mathbb{R}^{C_0 \times L}$, where $C_0$ denotes the base channel width.

Four successive residual blocks then progressively encode the representation. The first three blocks each apply a strided convolution with stride~2, yielding intermediate feature maps $\mathbf{F}_1 \in \mathbb{R}^{C_1 \times L/2}$, $\mathbf{F}_2 \in \mathbb{R}^{C \times L/4}$, and $\mathbf{F}_3 \in \mathbb{R}^{C \times L'}$, where $C_1 = 2C_0$, $C = 4C_0$, and $L' = L/8$. The fourth block operates at the same spectral resolution, producing $\mathbf{F}_4 \in \mathbb{R}^{C \times L'}$. Each residual block comprises two convolutional layers, each followed by batch normalization, with a GELU activation applied after the first convolution--batch normalization (Conv-BN) layer and again after the residual addition, together with a squeeze-and-excitation (SE) module for adaptive channel-wise recalibration. Formally, for block $i \in \{1,2,3,4\}$ with input $\mathbf{F}_{i-1}$ (where $\mathbf{F}_0 = \mathbf{H}_0$), the residual branch is defined as:

\begin{equation}
    f_i(\mathbf{F}) = \text{BN}\!\left(\mathbf{W}_i^{(2)} \ast \text{GELU}\!\left(\text{BN}\!\left(\mathbf{W}_i^{(1)} \ast \mathbf{F}\right)\right)\right),
\end{equation}

where $\mathbf{W}_i^{(1)}$ and $\mathbf{W}_i^{(2)}$ are the convolutional filters of the two Conv-BN layers within block~$i$. The block output is then obtained by adding the residual branch to the shortcut and applying a final activation:

\begin{equation}
    \tilde{\mathbf{F}}_i = \text{GELU}\!\left(f_i(\mathbf{F}_{i-1}) + \mathcal{P}_i(\mathbf{F}_{i-1})\right),
\end{equation}

where $\mathcal{P}_i(\cdot)$ is a strided pointwise projection for the first three blocks and the identity mapping for the fourth. The SE module then computes a channel-wise attention vector:

\begin{equation}
    \mathbf{s}_i = \sigma\!\left(\mathbf{W}_{i,2}^{\text{se}}\,\text{ReLU}\!\left(\mathbf{W}_{i,1}^{\text{se}}\,\text{GAP}(\tilde{\mathbf{F}}_i)\right)\right),
\end{equation}

where $\text{ReLU}(\cdot)$ denotes the rectified linear unit activation, $\text{GAP}(\cdot)$ denotes global average pooling, and $\mathbf{W}_{i,1}^{\text{se}}$ and $\mathbf{W}_{i,2}^{\text{se}}$ are pointwise linear projections. The recalibrated block output is then $\mathbf{F}_i = \tilde{\mathbf{F}}_i \odot \mathbf{s}_i$, where $\mathbf{s}_i$ is broadcast along the spectral dimension.

To preserve fine-grained spectral details from shallower layers, multi-scale feature fusion is performed by resampling $\mathbf{F}_2$ and $\mathbf{F}_1$ to resolution $L'$ via nearest-neighbor interpolation $\mathcal{R}(\cdot)$, with $\mathbf{F}_1$ additionally projected to $C$ channels via a pointwise convolution $\phi(\cdot)$. The resampled features are concatenated with $\mathbf{F}_4$ along the channel dimension to form

\begin{equation}
    \mathbf{Y} =
    \bigl[\mathbf{F}_4,\;\mathcal{R}(\mathbf{F}_2),\;\phi\!\left(\mathcal{R}(\mathbf{F}_1)\right)\bigr]
    \in \mathbb{R}^{3C \times L'} .
\end{equation}

A pointwise channel-mixing multilayer perceptron (MLP) $\psi(\cdot)$ is then applied independently at each spectral position to reduce the channel dimension from $3C$ to $C$, producing the unified feature map:

\begin{equation}
    \mathbf{F} = \psi(\mathbf{Y}) \in \mathbb{R}^{C \times L'} .
\end{equation}

\subsubsection*{Patch-based Transformer encoder.}
The convolutional feature map $\mathbf{F} \in \mathbb{R}^{C \times L'}$ is tokenized by a strided convolutional patch embedding layer with kernel size $P$ and stride $P/2$, which projects each overlapping local segment into a $d$-dimensional token embedding and produces a sequence of $N$ patch tokens $\mathbf{E} \in \mathbb{R}^{N \times d}$. A learnable \texttt{[CLS]} token $\mathbf{e}_{\text{cls}} \in \mathbb{R}^{d}$ is prepended, and a learnable positional embedding $\mathbf{E}_{\text{pos}} \in \mathbb{R}^{(N+1) \times d}$ is added to encode sequential positional information, yielding the initial token sequence:

\begin{equation}
    \mathbf{T}^{(0)} = \bigl[\mathbf{e}_{\text{cls}};\,\mathbf{E}\bigr] + \mathbf{E}_{\text{pos}} \in \mathbb{R}^{(N+1) \times d}.
\end{equation}

The sequence is then processed by $N_{\text{layers}}$ stacked Transformer encoder layers with pre-layer normalization, where each layer applies a multi-head self-attention sub-layer ($H$ heads, per-head dimension $d_h = d/H$) followed by a two-layer feed-forward network with GELU activation and hidden dimension $4d$, each preceded by layer normalization and wrapped with a residual connection. The final hidden state corresponding to the \texttt{[CLS]} token is extracted from the last Transformer layer as the global latent spectral representation:

\begin{equation}
    \mathbf{z} = \mathbf{T}^{(N_{\text{layers}})}_{[0]} \in \mathbb{R}^{d}.
\end{equation}

\subsection*{Task-specific downstream architectures}

The hybrid convolutional--Transformer encoder is shared across downstream tasks. Most classification and regression tasks apply the shared spectral encoder to a single input spectrum and pass the resulting representation to a task-specific MLP head. Two task families require additional task-specific modules. Formula-conditioned molecular structure elucidation takes a molecular formula as an additional input and generates SMILES strings. Pairwise mixture analysis takes a reference single-compound spectrum together with a mixture spectrum to determine whether the reference compound is present and to estimate its fractional contribution. By contrast, mixture-level component quantification from a mixture spectrum alone follows the standard single-spectrum regression pipeline. The following sections describe the additional modules used for structure generation and pairwise mixture analysis.

\subsubsection*{Molecular structure elucidation.}

Molecular structure elucidation is formulated as formula-conditioned autoregressive SMILES generation. Given an IR spectrum and its associated molecular formula, the model generates a ranked list of candidate SMILES strings by connecting the pretrained encoder to a molecular formula encoder, a Perceiver-style IR token resampler, a spectral vocabulary prompt module, and an autoregressive SMILES decoder.

The molecular formula string is tokenized at the character level and encoded by a formula encoder. The encoder consists of a token embedding layer with sinusoidal positional encodings and dropout, followed by $N_f$ pre-norm Transformer encoder layers, each with multi-head self-attention and a GELU feed-forward sublayer, and a final layer normalization. It produces formula hidden states $\mathbf{H}_f \in \mathbb{R}^{M_f \times d}$, where $M_f$ denotes the number of formula tokens.

On the spectral side, the backbone extracts both the global \texttt{[CLS]} feature $\mathbf{z} \in \mathbb{R}^{d}$ and the full patch token sequence $\mathbf{T}_{\mathrm{patch}} \in \mathbb{R}^{N \times d}$ from its final Transformer layer. A projection head comprising layer normalization, a linear layer, and GELU activation maps $\mathbf{z}$ to a single IR memory token $\mathbf{m}_{\mathrm{IR}} \in \mathbb{R}^{d}$. The patch tokens are further compressed by a Perceiver-style resampler into $K$ latent vectors. The resampler maintains $K$ learnable query embeddings and refines them over $N_r$ cross-attention blocks. Unlike standard post-norm Perceivers, each block applies pre-norm layer normalization separately to the queries and the key--value source before cross-attention. A pre-norm GELU feed-forward sublayer follows, with residual connections throughout. Denoting the latent state at layer $l$ as $\mathbf{X}^{(l)} \in \mathbb{R}^{K \times d}$, the update rule per block is:

\begin{align}
    \mathbf{X}^{(l)} &= \mathbf{X}^{(l-1)} + \mathrm{CrossAttn}\!\left(\mathrm{LN}_{q}\!\left(\mathbf{X}^{(l-1)}\right),\;\mathrm{LN}_{kv}\!\left(\mathbf{T}_{\mathrm{patch}}\right)\right), \\
    \mathbf{X}^{(l)} &\leftarrow \mathbf{X}^{(l)} + \mathrm{FFN}\!\left(\mathbf{X}^{(l)}\right),
\end{align}

with $\mathbf{X}^{(0)} = \mathbf{Q}_{\mathrm{lat}} \in \mathbb{R}^{K \times d}$ the learnable query matrix, yielding $\mathbf{M}_{\mathrm{lat}} = \mathbf{X}^{(N_r)} \in \mathbb{R}^{K \times d}$.

A soft vocabulary prompt token is constructed from $\mathbf{z}$ to bias the decoder toward chemically plausible token sequences. A task-specific MLP head maps $\mathbf{z}$ to logits over the SMILES vocabulary of size $V$. A temperature-scaled softmax then converts these logits into a probability distribution, which is used to compute a soft embedding as a weighted sum over the decoder embedding matrix $\mathbf{E} \in \mathbb{R}^{V \times d}$. The soft embedding is then projected and scaled by a learnable scalar gate $\alpha$ to yield the prompt token:

\begin{equation}
    \mathbf{p} = \mathrm{softmax}\!\left(\frac{\mathrm{MLP}_{\mathrm{cls}}(\mathbf{z})}{\tau}\right) \in \mathbb{R}^{V},
    \qquad
    \mathbf{m}_{\mathrm{prompt}} = \alpha \cdot f_{\mathrm{proj}}\!\left(\mathbf{p}^{\top}\mathbf{E}\right) \in \mathbb{R}^{d},
\end{equation}

where $\mathrm{MLP}_{\mathrm{cls}}$ denotes the task-specific vocabulary-classification head, $\tau$ is a fixed temperature hyperparameter, and $f_{\mathrm{proj}}$ denotes a three-stage projection comprising layer normalization, a linear transformation, and GELU activation.

The complete decoder memory is formed by concatenating all spectral tokens and the formula hidden states:

\begin{equation}
    \mathbf{M} = \bigl[\mathbf{m}_{\mathrm{IR}};\;\mathbf{M}_{\mathrm{lat}};\;\mathbf{m}_{\mathrm{prompt}};\;\mathbf{H}_f\bigr]
    \in \mathbb{R}^{(K+2+M_f)\times d}.
\end{equation}

The SMILES decoder consists of $N_s$ pre-norm Transformer layers, each applying causal self-attention under a triangular mask, cross-attention to $\mathbf{M}$, and a GELU feed-forward sublayer, followed by a final layer normalization and a linear projection to vocabulary logits. During training, teacher forcing is applied, and the primary loss is token-level cross-entropy $\mathcal{L}_{\mathrm{LM}}$. Two contrastive losses further shape the representation. An IR--formula contrastive loss $\mathcal{L}_{\mathrm{IR\text{-}F}}$ aligns $\mathbf{z}$ with the masked mean-pooled formula encoding of $\mathbf{H}_f$, whereas an IR--target contrastive loss $\mathcal{L}_{\mathrm{IR\text{-}T}}$ aligns $\mathbf{z}$ with the masked mean-pooled decoder hidden states at the final Transformer layer. Both use symmetric information noise-contrastive estimation (InfoNCE) with two-layer projection heads that map the inputs into a shared $d_c$-dimensional contrastive space. The training objective is:

\begin{equation}
    \mathcal{L}_{\mathrm{struct}} = \mathcal{L}_{\mathrm{LM}}
    + \lambda_{\mathrm{IF}}\,\mathcal{L}_{\mathrm{IR\text{-}F}}
    + \lambda_{\mathrm{IT}}\,\mathcal{L}_{\mathrm{IR\text{-}T}}.
\end{equation}

At inference, beam search with length-normalized scoring generates multiple ranked candidate SMILES sequences. A formula-consistency reranking step then promotes candidates whose SMILES match the exact atomic composition of the provided molecular formula and demotes chemically invalid structures.

\subsubsection*{Targeted component detection and contribution estimation.}

The pairwise mixture-analysis tasks, targeted component detection and targeted fractional contribution estimation, are unified under a common spectral comparison framework. Given a reference single-compound spectrum and a mixture spectrum, the model predicts whether the reference compound is present in the mixture and estimates its fractional contribution. The two tasks share the same pairwise representation backbone and differ only in their final prediction head.

Each input pair $(\mathbf{x}_{\mathrm{ref}},\,\mathbf{x}_{\mathrm{mix}}) \in \mathbb{R}^{1\times L}\times\mathbb{R}^{1\times L}$ is encoded by three parallel branches that extract complementary representations of the spectral pair.

The first branch uses a weight-shared encoder to process each spectrum independently, producing a global \texttt{[CLS]} feature and a patch token sequence for each spectrum. The patch tokens are projected to $d_p$ dimensions and aggregated by attentive pooling, combining an attention-weighted sum with an element-wise maximum:

\begin{equation}
\mathbf{p}_{\mathrm{pool}} = \tfrac{1}{2}\!\left(\sum_{n=1}^{N}\alpha_n\,\mathbf{t}_n + \operatorname*{max}_{n}\,\mathbf{t}_n\right), \qquad \alpha_n = \frac{\exp(s_n)}{\sum_{n'}\exp(s_{n'})},
\end{equation}

where $\mathbf{t}_n$ are the projected patch tokens and $s_n$ is an attention score produced by layer normalization followed by a linear projection.

The second branch operates directly on the raw input signals. The reference spectrum, mixture spectrum, and their absolute pointwise difference are stacked into a three-channel tensor $[\mathbf{x}_{\mathrm{ref}},\,\mathbf{x}_{\mathrm{mix}},\,|\mathbf{x}_{\mathrm{mix}}-\mathbf{x}_{\mathrm{ref}}|] \in \mathbb{R}^{3\times L}$. Successive Conv-BN-GELU blocks with progressively doubling channel widths process this tensor and reduce its spectral resolution by a fixed factor. The resulting feature sequence is augmented with sinusoidal positional encodings, processed by a shallow Transformer encoder, attentively pooled, and projected to yield the joint feature $\mathbf{h}_{\mathrm{joint}} \in \mathbb{R}^{d_p}$.

The third branch provides complementary frequency-domain features. A real-input fast Fourier transform (FFT) is applied to each spectrum, yielding complex coefficients $\hat{r}_f$ and $\hat{m}_f$ for the reference and mixture spectra at frequency bin $f$. Eight frequency-domain channels are then constructed: the log-amplitude spectra $\log(1+|\hat{r}_f|)$ and $\log(1+|\hat{m}_f|)$, their sum and absolute difference, the normalized cross-magnitude term $\kappa_f = |\hat{r}_f\hat{m}_f^{*}|/(|\hat{r}_f|\,|\hat{m}_f|+\varepsilon)$, the normalized cross-spectral phase $\phi_f = \angle(\hat{r}_f\hat{m}_f^{*})/\pi$, the log-amplitude ratio $\rho_f=\log\!\left(1+(|\hat{m}_f|+\varepsilon)/(|\hat{r}_f|+\varepsilon)\right)$, and the log-amplitude product $\pi_f=\log\!\left(1+|\hat{r}_f|\,|\hat{m}_f|\right)$, where $(\cdot)^*$ denotes complex conjugation, $\angle(\cdot)$ denotes the complex argument, and $\varepsilon$ is a small constant for numerical stability. These channels are concatenated to form the frequency-domain tensor $\mathbf{X}_{\mathrm{freq}}\in\mathbb{R}^{8\times F}$, which is processed by three convolutional sub-branches with short, medium, and long receptive fields. The resulting feature maps are concatenated, compressed by strided convolutions, globally average-pooled, and projected to yield the frequency representation $\mathbf{h}_{\mathrm{freq}}\in\mathbb{R}^{d_p}$.

The pair representation $\mathbf{r}_{\mathrm{pair}}$ is assembled by concatenating five $d_p$-dimensional feature vectors with a ten-dimensional scalar statistics vector $\mathbf{s}_{\mathrm{stats}}$. The five feature vectors are the projected \texttt{[CLS]} features $\mathbf{c}_{\mathrm{ref}}$ and $\mathbf{c}_{\mathrm{mix}}$, their absolute difference $|\mathbf{c}_{\mathrm{ref}}-\mathbf{c}_{\mathrm{mix}}|$, the absolute difference of the pooled token features $|\mathbf{p}_{\mathrm{ref}}-\mathbf{p}_{\mathrm{mix}}|$, and the frequency feature $\mathbf{h}_{\mathrm{freq}}$. The statistics vector $\mathbf{s}_{\mathrm{stats}}$ collects the cosine similarity between the \texttt{[CLS]} features, the cosine similarity between the pooled-token features, the cosine similarity between $\mathbf{h}_{\mathrm{joint}}$ and $\mathbf{h}_{\mathrm{freq}}$, the mean squared activation energy of the joint Transformer tokens, the cosine similarity of the log-amplitude spectra, the mean and maximum absolute log-amplitude difference, the mean and maximum normalized cross-magnitude term, and the mean absolute normalized cross-spectral phase:

\begin{equation}
\mathbf{r}_{\mathrm{pair}} = \bigl[\mathbf{c}_{\mathrm{ref}};\;\mathbf{c}_{\mathrm{mix}};\;|\mathbf{c}_{\mathrm{ref}}-\mathbf{c}_{\mathrm{mix}}|;\;|\mathbf{p}_{\mathrm{ref}}-\mathbf{p}_{\mathrm{mix}}|;\;\mathbf{h}_{\mathrm{freq}};\;\mathbf{s}_{\mathrm{stats}}\bigr].
\end{equation}

The pair representation $\mathbf{r}_{\mathrm{pair}}$ is then passed to a task-specific prediction head. For targeted component detection, a binary classification head maps $\mathbf{r}_{\mathrm{pair}}$ to a scalar logit and is trained with binary cross-entropy loss. For targeted fractional contribution estimation, a regression head maps $\mathbf{r}_{\mathrm{pair}}$ to a scalar contribution estimate $\hat{a}\in[0, 1]$ through a sigmoid output layer and is trained with mean squared error against the normalized target contribution.

\subsection*{Pretraining}

In the large-scale simulation-based pretraining stage, each preprocessed IR spectrum $\mathbf{x}$ is first stochastically augmented to obtain an encoder input $\tilde{\mathbf{x}}$. The augmented spectrum is passed through the hybrid convolutional--Transformer encoder $f_{\theta}$, producing the latent representation $\mathbf{z}=f_{\theta}(\tilde{\mathbf{x}})$. Three complementary objectives are then applied to outputs derived from $\mathbf{z}$: wavelet-domain spectral reconstruction, molecular fingerprint similarity alignment, and multi-label functional-group prediction. Together, these objectives encourage the learned representation to capture spectral morphology, molecular structural relationships, and interpretable functional-group signatures. The total pretraining loss is their weighted combination:

\begin{equation}
    \mathcal{L}_{\text{pre}} =
    \lambda_{\text{recon}}\,\mathcal{L}_{\text{recon}}
    + \lambda_{\text{contrast}}\,\mathcal{L}_{\text{contrast}}
    + \lambda_{\text{fg}}\,\mathcal{L}_{\text{fg}},
\end{equation}

where $\lambda_{\text{recon}}$, $\lambda_{\text{contrast}}$, and $\lambda_{\text{fg}}$ are scalar loss weights. The encoder and pretraining heads are optimized jointly by backpropagating $\mathcal{L}_{\text{pre}}$ using AdamW. Each objective is described below.

\subsubsection*{Wavelet-domain spectral reconstruction.}

To encourage the encoder to preserve fine-grained spectral structure, wavelet-domain spectral reconstruction is used as one of the pretraining objectives. A dedicated wavelet reconstruction head maps $\mathbf{z}$ through a two-stage bottleneck, each stage comprising layer normalization, a linear transformation, a GELU activation, and dropout, yielding a compressed representation $\mathbf{b} \in \mathbb{R}^{d_b}$. This representation is decoded into the approximation coefficients and $J=4$ levels of detail coefficients of a discrete wavelet transform (DWT) using a Daubechies-4 wavelet. Specifically, the predicted approximation coefficients $\hat{\mathbf{a}} \in \mathbb{R}^{L_0}$ and the $j$-th-level detail coefficients $\hat{\mathbf{d}}_j \in \mathbb{R}^{L_j}$ are given by:

\begin{equation}
    \hat{\mathbf{a}} = e^{\alpha_0}\,\mathbf{W}_{a}\,\mathbf{b},
    \qquad
    \hat{\mathbf{d}}_j = e^{\alpha_j}\,\mathbf{W}_{d,j}\,\mathbf{b}, \quad j = 1,\ldots,J,
\end{equation}

where $\mathbf{W}_{a}$ and each $\mathbf{W}_{d,j}$ are linear projections to the respective band lengths, and $\alpha_0, \alpha_1, \ldots, \alpha_J$ are learnable log-scale parameters that explicitly accommodate the heterogeneous energy scales across wavelet subbands. The predicted coefficients are then passed through the inverse DWT (IDWT) to recover the full-resolution spectral estimate $\hat{\mathbf{x}} \in \mathbb{R}^{L}$.

The reconstruction objective combines a wavelet-coefficient-domain loss and a signal-domain loss. Denoting by $\mathbf{a}$ and $\{\mathbf{d}_j\}_{j=1}^{J}$ the target DWT coefficients of the unaugmented spectrum $\mathbf{x}$, the coefficient loss applies the smooth-$\ell_1$ loss $\ell_{\mathrm{H}}(\cdot,\cdot)$ independently to each wavelet band and averages uniformly across all $J+1$ bands:

\begin{equation}
    \mathcal{L}_{\text{coeff}} = \frac{1}{J+1}\left[\ell_{\mathrm{H}}\!\left(\hat{\mathbf{a}},\,\mathbf{a}\right) + \sum_{j=1}^{J}\ell_{\mathrm{H}}\!\left(\hat{\mathbf{d}}_j,\,\mathbf{d}_j\right)\right].
\end{equation}

The signal-domain loss is computed between $\hat{\mathbf{x}}$ and the ground-truth $\mathbf{x}$ using a spatially weighted smooth-$\ell_1$ loss. Leveraging the binary mask $\mathbf{m}$ introduced during data augmentation, masked positions are assigned an elevated weight $m_w > 1$ to concentrate reconstruction capacity on the corrupted spectral regions. The per-position weight is $w_l = 1 + (m_w - 1)\,m_l$, and the signal-domain loss is computed as the weighted average:

\begin{equation}
    \mathcal{L}_{\text{signal}} =
    \frac{
        \sum_{l=1}^{L} w_l\;\ell_{\mathrm{H}}\!\left(\hat{x}_l,\,x_l\right)
    }{
        \sum_{l=1}^{L} w_l
    }.
\end{equation}

The two reconstruction terms are combined as a weighted sum:

\begin{equation}
    \mathcal{L}_{\text{recon}} =
    \lambda_{\text{coeff}}\,\mathcal{L}_{\text{coeff}}
    + \lambda_{\text{signal}}\,\mathcal{L}_{\text{signal}}.
\end{equation}

By jointly supervising coarse-scale spectral envelopes through the approximation band and fine-scale absorption line shapes through the detail bands, this pretraining objective encourages the encoder to learn representations that capture both global spectral morphology and local absorption features.

\subsubsection*{Molecular fingerprint similarity alignment.}

To encourage the latent space to reflect graded structural relationships among molecules, UltraIR is additionally trained with a soft Tanimoto contrastive objective. This objective operates on a dedicated projection of the encoder output: a two-layer projection head maps $\mathbf{z}$ to a $d_p$-dimensional embedding $\mathbf{p}\in\mathbb{R}^{d_p}$ via layer normalization, a linear projection, a GELU activation, dropout, and a final linear projection.

Rather than defining hard positive and negative pairs, this objective constructs a continuous structural similarity target from precomputed binary molecular fingerprints. For a batch of $B$ samples with binarized fingerprints $\mathbf{f}_i \in \{0,1\}^{M}$, the pairwise Tanimoto similarity is:

\begin{equation}
    T_{ij} =
    \frac{\mathbf{f}_i^{\top}\mathbf{f}_j}
    {\|\mathbf{f}_i\|_1 + \|\mathbf{f}_j\|_1 - \mathbf{f}_i^{\top}\mathbf{f}_j + \varepsilon}.
\end{equation}

A soft target distribution $p_{ij}$ over off-diagonal batch members is obtained by applying a softmax with temperature $\tau_t$ to each row of $T$ after masking the diagonal. Student similarity logits are computed as scaled inner products between $\ell_2$-normalized embeddings $\bar{\mathbf{p}}_i=\mathbf{p}_i/\|\mathbf{p}_i\|_2$, giving $s_{ij}=\bar{\mathbf{p}}_i^{\top}\bar{\mathbf{p}}_j/\tau_s$. The contrastive loss is the soft cross-entropy between the student distribution and the Tanimoto-derived target:

\begin{equation}
    \mathcal{L}_{\text{contrast}} =
    -\frac{1}{B}
    \sum_i
    \sum_{j\neq i}
    p_{ij}
    \log
    \frac{\exp(s_{ij})}
    {\sum_{k\neq i}\exp(s_{ik})}.
\end{equation}

This objective encourages spectra of structurally related compounds to occupy nearby regions of the latent space in proportion to their fingerprint overlap.

\subsubsection*{Multi-label functional-group prediction.}

A multilayer classification head maps the latent representation $\mathbf{z}$ to logits over the functional-group classes. It consists of layer normalization followed by three linear projections with progressively reducing dimensionality, interleaved with GELU activations and dropout. The model is trained to predict the presence or absence of each functional group as an independent binary classification task using binary cross-entropy with logits. This objective provides explicit chemical supervision that anchors the latent space to interpretable structural motifs.

\subsection*{Downstream adaptation}

For downstream adaptation, the pretrained checkpoint was used to initialize only the shared spectral encoder components present in both the pretraining and downstream models. These components included the derivative-aware multi-channel input module, the convolutional spectral encoder, and the patch-based Transformer encoder. Pretraining-specific heads, including the reconstruction head, contrastive projection head, and pretraining functional-group prediction head, were not transferred to downstream tasks. Consequently, even when a downstream task shared a prediction type or label space with a pretraining objective, its task-specific output module was newly initialized rather than inherited from the pretraining checkpoint. Other components introduced exclusively for downstream prediction were also initialized randomly.

\subsection*{Downstream task heads}

\subsubsection*{Classification head.}

All classification tasks use task-specific instances of the same MLP head architecture. The head consists of layer normalization followed by three linear projections with progressively reducing dimensionality, interleaved with GELU activations and a dropout layer. The final projection maps the hidden representation to $C$ output logits, where $C$ denotes the number of classes for each task. Single-spectrum classification tasks receive the global spectral representation $\mathbf{z}$ as input, whereas targeted component detection receives the dedicated pair representation $\mathbf{r}_{\mathrm{pair}}$.

For functional-group prediction, each output logit corresponds to an independent binary classification target, and the head is trained with binary cross-entropy loss. Bacterial classification, medicinal-herb geographic origin traceability, and microplastics classification are formulated as single-label multiclass classification tasks and trained with cross-entropy loss, with label smoothing applied for medicinal-herb geographic origin traceability. Targeted component detection produces a scalar logit and is trained with binary cross-entropy loss.

\subsubsection*{Regression head.}

For regression tasks, including physicochemical property prediction, targeted fractional contribution estimation, mixture-level component quantification, medicinal-herb constituent quantification, and soil property prediction, a task-specific three-layer MLP regression head processes the input representation. This head follows the same architecture as the classification head but uses task-specific output layers. Single-spectrum regression tasks receive the global spectral representation $\mathbf{z}\in\mathbb{R}^{d}$ as input, whereas targeted fractional contribution estimation receives the dedicated pair representation $\mathbf{r}_{\mathrm{pair}}$.

To stabilize training across target dimensions with heterogeneous numerical scales, regression targets were transformed using target-wise scaling. For single-target regression tasks, including targeted fractional contribution estimation, the final output is passed through a sigmoid activation and optimized using mean squared error loss. For multi-target regression tasks, including physicochemical property prediction, mixture-level component quantification, medicinal-herb constituent quantification, and soil property prediction, the regression head adopts a linear output layer and is optimized using a smooth-\(\ell_1\) regression loss.

\section{Data and code availability}
\label{sec:data_availability}

The code is available at \url{https://github.com/AIMS-Lab-HKUSTGZ/UltraIR}. The checkpoints of UltraIR are available at \url{https://huggingface.co/yusentan/UltraIR}. 

The newly generated UltraIR molecular-dynamics simulated infrared dataset used for UltraIR pretraining can be accessed at \url{https://huggingface.co/yusentan/UltraIR}. The simulated infrared dataset from IRtoMol \citep{irtomol} used for UltraIR pretraining can be accessed through \url{https://zenodo.org/records/7928396}.  
The simulated infrared dataset from the multimodal spectroscopy dataset \citep{nipsdata} used for UltraIR pretraining can be accessed through \url{https://zenodo.org/records/14770232}.  
The simulated infrared dataset from QM9S \citep{qm9s} used for UltraIR pretraining can be accessed through \url{https://figshare.com/articles/dataset/QM9S_dataset/24235333}.

The real infrared dataset from the NIST Chemistry WebBook \citep{nistwebbook} used for molecular-level interpretation and mixture analysis benchmarks can be accessed through \url{https://webbook.nist.gov/chemistry/}. The real infrared dataset from SDBS \citep{sdbs} used for molecular-level interpretation and mixture analysis benchmarks can be accessed through \url{https://sdbs.db.aist.go.jp/}. The simulated infrared dataset from USPTO \citep{USPTO} used for molecular-level interpretation and mixture analysis benchmarks can be accessed through \url{https://zenodo.org/records/16417648}. The external liquid-mixture benchmark from DeepMIR \citep{deepmir} used for mixture analysis benchmarks can be accessed through \url{https://github.com/LinTan-CSU/DeepMIR}. The experimental FTIR mixture dataset \citep{digital_discovery_2022} used for mixture analysis benchmarks can be accessed through \url{https://doi.org/10.5281/zenodo.5498197}.

The green-snow bacterial FTIR dataset \citep{bacteria} used for bacterial classification can be accessed through \url{https://zenodo.org/records/4297950}. The two newly generated in-house experimental datasets of Jinyinhua and Shanyinhua used for medicinal-herb characterization can be accessed at \url{https://huggingface.co/yusentan/UltraIR}. The environmentally sourced microplastics infrared dataset \citep{microp} used for microplastics classification can be accessed through \url{https://drive.google.com/drive/folders/11MofhjEchgZelWPcHUvIMRPNEQPaLfUO?usp=sharing}. The OSSL dataset \citep{ossl} used for soil property prediction can be accessed through \url{https://docs.soilspectroscopy.org/db-access.html}.
\section{Acknowledgments}
\label{sec:acknowledgements}

This research was supported by the National Natural Science Foundation of China Project (No. 623B2086), CCF-GHFund (No. OF 2026005), the CIPS-SMP-Zhipu Large Model Fund, Ant Group, Shanghai Artificial Intelligence Laboratory, and TeleAI of China Telecom.

\section{Author contributions}
\label{sec:author_contributions}

Conceptualization, Y.T. and J.X.; methodology, Y.T., Y.C., and J.X.; software, Y.T.; data curation, Y.T., Y.C., Z.F., P.L., and Y.F.L.; formal analysis, Y.T., Y.C., and Z.F.; investigation, Y.T., Y.C., Z.F., P.L., and Y.F.L.; validation, Y.T., Y.C., Z.F., P.L., and Y.F.L.; visualization, Y.T., Y.C., Q.G., and Z.L.; writing -- original draft, Y.T.; writing -- review \& editing, all authors; resources, J.X., Y.Q.L., and T.W.; supervision, J.X., X.Z., and T.W.; project administration, J.X.; funding acquisition, J.X.

\section{Competing interests}
\label{sec:competing_interests}

The authors declare no competing interests.

\bibliographystyle{unsrtnat}
\bibliography{ref}

\clearpage
\beginappendix

\renewcommand\thefigure{\Alph{figure}}
\renewcommand\thetable{\Alph{table}}
\renewcommand\thesection{\Alph{section}}
\setcounter{figure}{0}
\setcounter{table}{0}
\setcounter{section}{0}

\section{Supplementary experimental details}
\label{supp:experimental_details}

\subsection*{Pretraining hyperparameters}
\label{supp:pretraining_hyperparameters}

The principal hyperparameters used for large-scale UltraIR pretraining are summarized in Table~\ref{tab:pretraining_hyperparameters}.

\begin{table}[!htb]
\centering
\caption{
\textbf{Pre-training hyperparameters of UltraIR.}
}
\label{tab:pretraining_hyperparameters}
\begin{tabular}{ll}
\hline
Hyperparameter & Value \\
\hline

\multicolumn{2}{l}{\textit{Optimization}}\\
Learning rate & $1\times10^{-4}$ \\
Batch size & 128 \\
Epochs & 5 \\
Warmup ratio & 5\% \\

\hline

\multicolumn{2}{l}{\textit{Transformer encoder}}\\
Hidden dimension \(d\) & 1024 \\
Patch length \(P\) & 16 \\
Number of attention heads \(H\) & 16 \\
Number of Transformer layers \(N_{\mathrm{layers}}\) & 8 \\
Dropout & 0.05 \\

\hline

\multicolumn{2}{l}{\textit{Pre-training heads}}\\
Fingerprint embedding dimension \(d_p\) & 512 \\
Number of wavelet decomposition levels \(J\) & 4 \\
Wavelet reconstruction bottleneck dimension \(d_b\) & 256 \\

\hline

\multicolumn{2}{l}{\textit{Pre-training objectives}}\\
Wavelet-domain spectral reconstruction loss weight
\(\lambda_{\mathrm{recon}}\) & 50.0 \\
Molecular fingerprint similarity alignment loss weight
\(\lambda_{\mathrm{contrast}}\) & 0.2 \\
Multi-label functional-group prediction loss weight
\(\lambda_{\mathrm{fg}}\) & 1.25 \\

Wavelet-coefficient-domain loss weight
\(\lambda_{\mathrm{coeff}}\) & 0.2 \\
Signal-domain loss weight
\(\lambda_{\mathrm{signal}}\) & 0.8 \\
Masked-position weighting factor \(m_w\) & 3.0 \\

Fingerprint student temperature \(\tau_s\) & 0.1 \\
Fingerprint target temperature \(\tau_t\) & 0.05 \\

\hline
\end{tabular}
\end{table}

\subsection*{Pretraining dataset construction}
\label{supp:pretraining_dataset}

The UltraIR pretraining dataset comprised 60,000,000 simulated infrared (IR) spectra assembled from three complementary sources: publicly released simulated spectra, spectra generated using molecular dynamics, and spectra predicted using a machine-learning model. After removing molecules overlapping with the National Institute of Standards and Technology (NIST) Chemistry WebBook, Spectral Database for Organic Compounds (SDBS), and USPTO datasets, the publicly released sources contributed 630,395 spectra from IRtoMol, 743,955 from the multimodal spectroscopy dataset, and 127,898 from QM9S, yielding 1,502,248 spectra in total \citep{irtomol,nipsdata,qm9s}. From structures in the PubChem-derived molecular collection distributed with the SimNMR-PubChem database \citep{nmrsolver} that were not represented in NIST, SDBS, or USPTO, we generated 7,534,293 spectra through molecular-dynamics simulation and predicted the remaining 50,963,459 spectra were predicted using Chemprop-IR \citep{chemprop}.

For the molecular-dynamics component, three-dimensional molecular conformers were generated from Simplified Molecular Input Line Entry System (SMILES) representations and subjected to geometry relaxation before simulation. Each molecule was simulated without periodic boundary conditions at 300 K using a Langevin middle integrator in OpenMM, with the Open Force Field Sage 2.2.1 and fixed atom-centered partial charges \citep{eastman2017openmm,boothroyd2023development}. Molecular dipole trajectories were calculated from the fixed partial charges and instantaneous atomic positions. After temporal mean removal and Hann windowing, the three Cartesian dipole components were transformed independently using a fast Fourier transform. The resulting power spectra were summed, weighted by wavenumber, interpolated onto the common wavenumber grid, and normalized to a maximum intensity of one. 

For the machine-learning-generated component, IR spectra were predicted using the two officially released pretrained Chemprop-IR checkpoints \citep{chemprop}. The two models were applied independently, and their predicted intensities were combined using an equally weighted arithmetic mean to obtain the final spectrum for each molecular structure.

Functional-group labels and molecular fingerprints used during pretraining were generated from molecular structures using RDKit. Functional-group labels were encoded as 17-dimensional multi-hot vectors, with each class assigned when the molecular graph matched its corresponding SMILES Arbitrary Target Specification (SMARTS) definition. Multiple classes could therefore be assigned to the same molecule. The functional-group classes, output order, and operational SMARTS definitions are provided in Table~\ref{tab:functional_group_smarts}. For molecular fingerprint similarity alignment, each structure was represented by a 2,048-bit Morgan fingerprint generated with a radius of 2.

\begin{table}[!tb]
\centering
\caption{\textbf{Functional-group classes and SMARTS definitions used for pretraining and downstream functional-group prediction.}
The listed order corresponds to the 17 dimensions of the multi-label target used in both stages. A class was assigned when the corresponding SMARTS pattern matched the molecular graph, and each molecule could receive multiple labels.}
\label{tab:functional_group_smarts}
\small
\begin{tabular}{c l p{0.34\textwidth}}
\toprule
Index & Functional-group class & SMARTS definition \\
\midrule
1  & Alkane
   & \texttt{[CX4;H3,H2]} \\

2  & Methyl
   & \texttt{[CH3]} \\

3  & Alkene
   & \texttt{[CX3]=[CX3]} \\

4  & Alkyne
   & \texttt{[CX2]\#[CX2]} \\

5  & Alcohols
   & \texttt{[\#6][OX2H]} \\

6  & Amines
   & \texttt{[NX3;H2,H1,H0;!\$(N[CX3](=O))]} \\

7  & Nitriles
   & \texttt{[NX1]\#[CX2]} \\

8  & Aromatics
   & \texttt{[\$([cX3](:*):*),\$([cX2+](:*):*)]} \\

9  & Alkyl halides
   & \texttt{[\#6][F,Cl,Br,I]} \\

10 & Esters
   & \texttt{[\#6][CX3](=O)[OX2H0][\#6]} \\

11 & Ketones
   & \texttt{[\#6][CX3](=O)[\#6]} \\

12 & Aldehydes
   & \texttt{[CX3H1](=O)[\#6]} \\

13 & Carboxylic acids
   & \texttt{[CX3](=O)[OX2H1]} \\

14 & Ether
   & \texttt{[OD2]([\#6;!\$(C=O)])([\#6;!\$(C=O)])} \\

15 & Acyl halides
   & \texttt{[CX3](=[OX1])[F,Cl,Br,I]} \\

16 & Amides
   & \texttt{[NX3][CX3](=[OX1])[\#6]} \\

17 & Nitro
   & \texttt{[\$([N+](=O)[O-]),\$([NX3](=O)=O)][\#6]} \\
\bottomrule
\end{tabular}
\end{table}

\subsection*{Downstream datasets and benchmark summary}
\label{supp:dataset_task_summary}

The eight downstream tasks were evaluated using ten datasets, with the Jinyinhua and Shanyinhua datasets each contributing separate subsets for geographic origin traceability and chemical constituent quantification. For the molecular-level benchmarks, we used NIST, SDBS, and USPTO. The NIST Chemistry WebBook, maintained as NIST Standard Reference Database 69, is a public resource that compiles chemical and spectroscopic data, including IR spectra \citep{nistwebbook}. We retained 23,286 NIST IR spectra after filtering. SDBS is a public database maintained by the National Institute of Advanced Industrial Science and Technology and contains experimentally measured Fourier-transform infrared (FTIR) spectra with corresponding molecular records for organic compounds \citep{sdbs}. We retained 18,253 SDBS IR spectra after filtering. The USPTO dataset is a computational multimodal spectroscopy resource constructed from patent-extracted organic molecules and contains 177,461 IR spectra generated using long-timescale molecular-dynamics simulations and machine-learning-assisted dipole-moment prediction \citep{USPTO}.

For functional-group prediction, labels for NIST, SDBS, and USPTO were generated using the same 17 classes, class order, and SMARTS definitions as in pretraining (Table~\ref{tab:functional_group_smarts}). For physicochemical property prediction, all 11 targets were calculated from molecular structures using RDKit. The synthetic accessibility score (SAScore) was obtained using the RDKit SA\_Score implementation, and the remaining targets were calculated using their corresponding RDKit descriptor functions.

Mixture-level and application-facing evaluations used several additional data sources. For external evaluation of targeted component detection, we used 360 spectra from the external liquid-mixture benchmark described in DeepMIR, comprising binary, ternary and quaternary mixtures \citep{deepmir}. Mixture-level component quantification was evaluated using 74 experimentally acquired spectra from a four-component FTIR mixture dataset comprising acrylonitrile, adiponitrile, propionitrile and glycerol \citep{digital_discovery_2022}. For this task, models were first trained on a simulated four-component mixture dataset containing 2,400 synthetic mixtures, and subsequently fine-tuned on the experimentally acquired spectra for evaluation \citep{digital_discovery_2022}. Bacterial genus classification used 795 FTIR spectra from 45 fast-growing bacterial isolates collected from Antarctic green snow and spanning nine genera \citep{bacteria}. For medicinal-herb geographic origin traceability and constituent quantification, the dataset composition, LC-MS-derived relative-abundance regression targets, together with the corresponding analytical procedures, are detailed in the following subsection. The microplastics benchmark contained 32,965 IR spectra assembled from laboratory and open-source collections to represent diverse environmentally relevant polymer classes \citep{microp}. After filtering for complete wavenumber coverage and availability of the labels required for soil property prediction, we retained 46,099 mid-IR spectra from the Kellogg Soil Survey Laboratory (KSSL) collection within the Open Soil Spectral Library (OSSL) \citep{ossl}. For cross-instrument and cross-laboratory evaluation, we independently filtered the datasets using the labels required for the cross-laboratory tasks and enforced complete wavenumber coverage. This procedure retained 55,974 KSSL spectra for model training and 3,753 ICRAF--ISRIC spectra for zero-shot target-domain inference. Table~\ref{tab:dataset_task_summary} summarizes the datasets and comparison methods used for each downstream task.

\begin{table}[!t]
\centering
\caption{
\textbf{Summary of downstream benchmarks and comparison methods.}
The table summarizes the datasets and competing methods evaluated for each downstream task.
}
\label{tab:dataset_task_summary}

\renewcommand{\arraystretch}{1.25}
\setcellgapes{2.5pt}
\makegapedcells

\resizebox{\textwidth}{!}{
\begin{tabular}{lll}
\hline
Task & Dataset & Compared methods \\
\hline

Functional-group prediction 
& NIST \cite{nistwebbook}, SDBS \cite{sdbs}, USPTO \cite{USPTO} 
& \makecell[l]{FCGFormer \cite{fg1}, IRAnalysis \cite{fg2}, XGBoost,\\ Random Forest, KNN, Logistic classifier} \\

Molecular structure elucidation 
& NIST \cite{nistwebbook}, SDBS \cite{sdbs}, USPTO \cite{USPTO} 
& IRtoMol \cite{irtomol}, AISE \cite{ibm}, PBSA \cite{denovo}, DLIR \cite{dlir}\\

Physicochemical property prediction 
& NIST \cite{nistwebbook}, SDBS \cite{sdbs}, USPTO \cite{USPTO} 
& XGBoost regression, SVR, KNN regression, PLSR\\

Targeted component detection 
& \makecell[l]{NIST \cite{nistwebbook}, SDBS \cite{sdbs}, USPTO \cite{USPTO} ,\\
DeepMIR liquid-mixture dataset \cite{deepmir}}
& DeepMIR \cite{deepmir}, reverse match, HQI\\

Targeted fractional contribution estimation 
& NIST \cite{nistwebbook}, SDBS \cite{sdbs}, USPTO \cite{USPTO} 
& \makecell[l]{ DeepMIR \cite{deepmir}, XGBoost regression,\\ SVR, KNN regression, PLSR}\\

Mixture-level component quantification 
& Experimental FTIR mixture dataset \cite{digital_discovery_2022}
& \makecell[l]{ AIMWSP \cite{acsnano_2023_inspired}, ML-FTIR \cite{digital_discovery_2022}, XGBoost regression,\\ SVR, KNN regression, PLSR}\\

Bacterial genus classification 
& Green-snow bacterial FTIR spectra \cite{bacteria}
& XGBoost, Random Forest, KNN, Logistic classifier\\

Medicinal-herb origin traceability 
& Jinyinhua, Shanyinhua 
& XGBoost, Random Forest, KNN, Logistic classifier\\

Medicinal-herb constituent quantification 
& Jinyinhua, Shanyinhua 
& XGBoost regression, SVR, KNN regression, PLSR\\

Microplastics classification 
& Environmentally sourced microplastics IR dataset \cite{microp}
& \makecell[l]{Softmax \cite{microp}, DB-CNN-CBAM \cite{dbcnn}, XGBoost,\\ Random Forest, KNN, Logistic classifier} \\

Soil property prediction 
& OSSL \cite{ossl}
& \makecell[l]{GADF-Swin \cite{gadfswin}, LSTM-CNN \cite{lstmcnn}, XGBoost regression,\\ SVR, KNN regression, PLSR}\\

\hline
\end{tabular}
}
\end{table}

\subsection*{In-house medicinal-herb datasets and LC-MS-derived regression targets}

In addition to the public and external benchmarks summarized above, we assembled two in-house medicinal-herb datasets: Jinyinhua (\textit{Lonicerae Japonicae Flos}) and Shanyinhua (\textit{Lonicerae Flos}). For geographic origin traceability, the Jinyinhua dataset contains 120 IR spectra from Shandong, Henan, Hebei, and Sichuan, and the Shanyinhua dataset contains 150 IR spectra from Hunan, Hubei, Sichuan, Henan, and Guangdong; each origin is represented by 30 spectra. For chemical constituent quantification, LC-MS-derived target relative abundances measured for 60 Jinyinhua and 75 Shanyinhua samples served as regression labels and were expressed in arbitrary units (a.u.). The Jinyinhua targets were 4-methylcoumarin, milrinone, geniposide, phenylacetaldehyde, cis-cinnamic acid, and lumazine; the Shanyinhua targets were coniferyl aldehyde, quercetin, kaempferol, and luteolin.

For LC-MS analysis, 30~mg of powdered material was extracted with 1.5~mL of 70\% aqueous methanol. The extract was vortexed, ultrasonicated at 30~$^\circ$C for 30~min, centrifuged at 12,000~r.p.m. for 15~min, and passed through a 0.22~$\mu$m membrane filter. Chromatographic separation was performed on a Kinetex F5 column (2.6~$\mu$m, 100~mm $\times$ 2.1~mm; Phenomenex) at 40~$^\circ$C. The mobile phases were 0.1\% formic acid in water (A) and 0.1\% formic acid in acetonitrile (B), with a flow rate of 0.2~mL~min$^{-1}$ and an injection volume of 4~$\mu$L. The percentage of A was programmed as 100\% at 0~min, 99\% at 2~min, 95\% at 3~min, 90\% at 6~min, 85\% at 14~min, 82\% at 15~min, 80\% at 18~min, 70\% at 20~min, 60\% at 23~min, 22\% at 31~min, 10\% at 33~min, 0\% at 36--44~min, and 100\% at 44.1--50~min. Full-scan time-of-flight mass spectrometry (TOF-MS) and information-dependent acquisition (IDA) tandem mass spectrometry (MS/MS) spectra were acquired in positive-ion mode over $m/z$ 100--1,000 and $m/z$ 50--1,000, respectively. The source temperature was 500~$^\circ$C, the ion-spray voltage was +5,500~V, ion-source gases 1 and 2 were each 50~psi, curtain gas was 30~psi, declustering potential was +80~V, and collision energy was 35~V.

\subsection*{Dataset metadata and label distributions for cross-instrument and cross-laboratory soil-property prediction}

For cross-instrument and cross-laboratory soil-property prediction, the source- and target-domain dataset metadata and corresponding label distributions across the nine soil properties are summarized in Tables~\ref{tab:crosslab_soil_dataset_metadata} and \ref{tab:crosslab_soil_label_distribution}, respectively.

\begin{table}[!tb]
\centering
\caption{\textbf{Dataset metadata for cross-instrument and cross-laboratory soil-property prediction.}
Spectra from the Kellogg Soil Survey Laboratory (KSSL) and ICRAF--ISRIC, together with their associated metadata, were obtained from the Open Soil Spectral Library (OSSL) \citep{ossl}.}
\label{tab:crosslab_soil_dataset_metadata}
\small
\renewcommand{\arraystretch}{1.15}
\begin{tabularx}{\textwidth}{
    l
    >{\raggedright\arraybackslash}X
    >{\raggedright\arraybackslash}X
}
\toprule
Characteristic &
\textbf{Kellogg Soil Survey Laboratory database}
\newline\textit{Source training domain} &
\textbf{ICRAF--ISRIC Soil Spectral Library}
\newline\textit{Target test domain} \\
\midrule
Institution
& USDA National Soil Survey Center
& World Agroforestry Centre / ISRIC \\

Sample origin
& United States
& Not recorded in OSSL \\

FTIR system
& Bruker Vertex 70 with HTS-XT
& Bruker Tensor 27 with HTS-XT \\

Sample preparation
& Ground to $<80$ mesh
& Ground to $<0.1$~mm \\

\bottomrule
\end{tabularx}
\end{table}

\begin{table}[!tb]
\centering
\caption{\textbf{Label distributions for cross-instrument and cross-laboratory soil-property prediction.}
The Kellogg Soil Survey Laboratory (KSSL) and ICRAF--ISRIC spectra and associated soil-property labels were obtained from the Open Soil Spectral Library (OSSL) \citep{ossl}. Values are reported as median [minimum, maximum].}
\label{tab:crosslab_soil_label_distribution}
\small
\renewcommand{\arraystretch}{1.15}
\begin{tabularx}{\textwidth}{
    l
    l
    >{\raggedright\arraybackslash}X
    >{\raggedright\arraybackslash}X
}
\toprule
Property & Unit &
\textbf{Kellogg Soil Survey Laboratory database}
\textit{Source training domain} &
\textbf{ICRAF--ISRIC Soil Spectral Library}
\newline\textit{Target test domain} \\
\midrule
Clay Content
& \%
& 20.88 [0, 96.14]
& 30.10 [0, 96.80] \\

Silt Content
& \%
& 39.10 [0, 94.50]
& 24.90 [0.20, 100.00] \\

Sand Content
& \%
& 32.60 [0.10, 100.00]
& 33.05 [0, 99.50] \\

pH (H$_2$O)
& --
& 6.37 [2.29, 10.70]
& 5.80 [3.00, 10.50] \\

CEC
& $\mathrm{cmol}_{\mathrm{c}}\,\mathrm{kg}^{-1}$
& 15.62 [0, 584.59]
& 11.50 [0, 189.60] \\

Exchangeable Ca
& $\mathrm{cmol}_{\mathrm{c}}\,\mathrm{kg}^{-1}$
& 11.92 [0, 410.41]
& 3.80 [0, 168.20] \\

Exchangeable Mg
& $\mathrm{cmol}_{\mathrm{c}}\,\mathrm{kg}^{-1}$
& 2.86 [0, 172.64]
& 1.10 [0, 68.00] \\

Exchangeable K
& $\mathrm{cmol}_{\mathrm{c}}\,\mathrm{kg}^{-1}$
& 0.364 [0, 32.33]
& 0.20 [0, 9.80] \\

Exchangeable Na
& $\mathrm{cmol}_{\mathrm{c}}\,\mathrm{kg}^{-1}$
& 0 [0, 868.36]
& 0.10 [0, 31.60] \\
\bottomrule
\end{tabularx}
\end{table}

\subsection*{Cross-validation and evaluation protocol}
\label{supp:experimental_protocol}

All downstream evaluations requiring supervised model training were conducted using five-fold cross-validation. For each such dataset, samples were partitioned into five non-overlapping folds, with each fold used once as the test set. In each cross-validation run, one fold, corresponding to approximately 20\% of the dataset, was held out for testing. The remaining samples were divided into training and validation subsets to approximate an overall 70:10:20 train--validation--test ratio. When this ratio could not be achieved exactly because of discrete sample counts, the partition with the smallest deviation from the target ratio was used. For the molecule-level NIST, SDBS, and USPTO benchmarks, scaffold-based partitioning was used to minimize structural overlap among the training, validation, and test sets. Identical partitions were used for UltraIR, the matched no-pretraining ablation where included, and all comparison methods, ensuring directly comparable evaluations.

For the reduced-training-data experiments, the validation and test sets remained fixed within each cross-validation run. Only the original training subset was randomly subsampled to the specified fractions of labeled training data. These fractions therefore refer to the available training subset rather than the complete dataset. At each fraction, identical subsampled training sets were used for all compared methods.

Model configurations, training duration, early-stopping criteria, checkpoint selection, and the optional use of downstream spectral augmentation were determined exclusively using the training and validation subsets within each cross-validation run. Weak physically motivated augmentations, when enabled, were applied only to training samples. Except for the DeepMIR liquid-mixture dataset, downstream performance was aggregated across the five held-out test folds. For DeepMIR, checkpoints trained on the corresponding NIST mixture-analysis tasks with matched target labels were applied directly for inference. No additional model training or five-fold cross-validation was performed on the DeepMIR dataset.

\subsection*{Evaluation metrics}
\label{supp:metric_conventions}

\subsubsection*{Statistical significance testing.}

All significance annotations shown in the figures were calculated independently for each dataset--metric combination. Methods were ranked according to their mean performance across the five held-out test folds, using the appropriate direction for each metric, and only the top-ranked and second-ranked methods were compared. Their fold-level results were paired by the corresponding test fold and analyzed using a two-sided paired \(t\)-test implemented with \texttt{scipy.stats.ttest\_rel}. The null hypothesis was that the mean paired difference between the two methods across the five folds was zero. Significance was annotated as **** for \(P < 0.0001\), *** for \(0.0001 \leq P < 0.001\), ** for \(0.001 \leq P < 0.01\), * for \(0.01 \leq P < 0.05\), and n.s. for \(P \geq 0.05\).

\subsubsection*{Classification metrics.}
Single-label classification tasks, including bacterial classification, medicinal-herb geographic origin traceability, and microplastics classification, were evaluated using accuracy, Macro-F1, and the multiclass Matthews correlation coefficient (MCC). Targeted component detection was evaluated using accuracy, Macro-F1, and the area under the receiver operating characteristic curve (ROC-AUC), calculated from the continuous component-presence scores before thresholding. Macro-F1 was calculated as the unweighted mean of the one-versus-rest F1 scores across classes, thereby assigning equal weight to each class. For the class-wise radar analyses of bacterial and microplastics classification, the value reported for each bacterial genus or polymer class was its corresponding one-versus-rest F1 score. These class-specific F1 scores were calculated independently for each class and were distinct from the Macro-F1 values reported for overall task performance.

Functional-group prediction was formulated as multi-label classification and evaluated using Micro-F1, Macro-F1, and exact match ratio (EMR). Micro-F1 was calculated by aggregating true positives, false positives, and false negatives across all samples and functional-group labels, whereas Macro-F1 was the unweighted mean of the label-specific F1 scores. EMR required the complete predicted functional-group vector to match the ground-truth vector:

\begin{equation}
\mathrm{EMR}
=
\frac{1}{N}
\sum_{i=1}^{N}
\mathbb{I}
\left(
{\mathbf{y}}_{i}
=
\hat{\mathbf{y}}_{i}
\right),
\end{equation}

where \(N\) is the number of evaluated spectra, \(\mathbf{y}_{i}\) and \(\hat{\mathbf{y}}_{i}\) are the ground-truth and predicted binary label vectors, respectively, and \(\mathbb{I}(\cdot)\) is the indicator function. A prediction containing either a missed functional group or an additional false-positive group was therefore counted as incorrect.

\subsubsection*{Molecular structure elucidation metrics.}
Formula-conditioned molecular structure elucidation was evaluated using top-\(k\) accuracy for \(k \in \{1,5,10\}\). Let \(s_i\) denote the ground-truth SMILES string for sample \(i\), and let \(\mathcal{P}_{i}^{(k)}\) denote the first \(k\) generated candidate SMILES strings in their original ranked order. Before structure matching, the ground-truth SMILES and every generated candidate were parsed using RDKit and converted to canonical isomeric SMILES. Denoting this canonicalization operation by \(\mathcal{C}(\cdot)\), top-\(k\) accuracy was defined as

\begin{equation}
\mathrm{Top}\mbox{-}k
=
\frac{1}{N}
\sum_{i=1}^{N}
\mathbb{I}
\left[
\exists\,
\hat{s}\in\mathcal{P}_{i}^{(k)}
\ \text{such that}\
\mathcal{C}(\hat{s})
=
\mathcal{C}(s_i)
\right].
\end{equation}

A sample was counted as correct when at least one of the first \(k\) generated candidates represented the same canonical molecular structure as the ground truth. Candidates that could not be parsed into valid molecular structures remained in their generated rank positions but could not contribute a correct match. Stereochemical annotations were retained during canonicalization where present in the corresponding SMILES strings.

Structural similarity between a generated candidate and the ground-truth molecule was assessed using the Tanimoto similarity of their binary molecular fingerprints. For fingerprint vectors \(\mathbf{f}\) and \(\mathbf{g}\), the Tanimoto similarity was defined as

\begin{equation}
T(\mathbf{f},\mathbf{g})
=
\frac{\mathbf{f}^{\mathsf{T}}\mathbf{g}}
{\lVert\mathbf{f}\rVert_{1}
+
\lVert\mathbf{g}\rVert_{1}
-
\mathbf{f}^{\mathsf{T}}\mathbf{g}}.
\end{equation}

The score ranges from \(0\) to \(1\), with larger values indicating greater fingerprint overlap.

\subsubsection*{Regression metrics.}
Regression tasks were evaluated using mean absolute error (MAE), root mean squared error (RMSE), and the coefficient of determination \(R^{2}\). For single-target regression tasks, including targeted fractional contribution estimation, MAE, RMSE, and \(R^{2}\) were calculated directly in the original target space.

For multi-target regression tasks, including physicochemical property prediction, mixture-level component quantification, medicinal-herb constituent quantification, and soil property prediction, normalized MAE and normalized RMSE were used to account for differences in numerical ranges among targets. These metrics were calculated after target-wise normalization and averaged across targets such that each target contributed equally. Overall \(R^{2}\) was calculated as the arithmetic mean of target-specific \(R^{2}\) values across targets.

For normalized residual analyses, residuals were calculated independently for each target and normalized by the corresponding target range. Let \(y_{ij}\) and \(\hat{y}_{ij}\) denote the ground-truth and predicted values, respectively, for target \(j\) of sample \(i\). The normalized residual was defined as

\begin{equation}
r'_{ij}
=
\frac{
\hat{y}_{ij}-y_{ij}
}{
y_{j,\max}-y_{j,\min}
},
\end{equation}

where \(y_{j,\max}\) and \(y_{j,\min}\) denote the maximum and minimum values of target \(j\), respectively. To visualize multi-target regression residuals, the normalized residuals from all samples and targets were pooled such that each sample--target pair contributed equally to the resulting distribution.

\subsubsection*{BertzCT complexity-thresholded analysis.}
Prediction errors across molecular complexity were assessed using cumulative BertzCT thresholds. For threshold \(\tau\), the evaluated subset was defined as

\begin{equation}
\mathcal{S}_{\tau}
=
\left\{
i \colon B_i \geq \tau
\right\},
\end{equation}

where \(B_i\) is the ground-truth BertzCT value of molecule \(i\). The mean relative error (MRE) was then calculated as

\begin{equation}
\mathrm{MRE}(\tau)
=
\frac{1}{|\mathcal{S}_{\tau}|}
\sum_{i\in\mathcal{S}_{\tau}}
\frac{
\left|
\hat{B}_{i}-B_{i}
\right|
}{
|B_i|+\epsilon
},
\end{equation}

where \(\epsilon\) is a small constant.

\subsubsection*{True-class margin analysis.}
For microplastics classification, prediction confidence was further examined using the true-class margin. Given the logit \(s_{ic}\) assigned by a model to class \(c\) for sample \(i\), with \(y_i\) denoting the ground-truth class label of sample \(i\), the margin was defined as

\begin{equation}
m_i
=
s_{i y_i}
-
\max_{c\neq y_i}
s_{ic}.
\end{equation}

A positive margin indicates that the ground-truth class received the highest logit, whereas a negative margin indicates that at least one competing class received a higher score. Larger positive values indicate greater separation between the true class and its closest competing class.

\newpage

\section{More results}

\subsection*{Extended ablation analysis of UltraIR}

\begin{table}[!htb]
\centering
\small
\setlength{\tabcolsep}{9pt}
\caption{
\textbf{Extended ablation analysis of simulated pretraining across representative downstream tasks.}
UltraIR is compared with the matched no-pretraining ablation across four downstream tasks. Values are means $\pm$ standard deviations across five test folds. Bold values indicate the better result for each metric. Arrows indicate the preferred direction.
}
\label{tab:extended_ablation}

\begin{tabular}{@{}lccc@{}}
    \toprule
    \toprule

    \multicolumn{4}{@{}l@{}}{
    \textbf{a, Functional-group prediction}
    \quad \textit{Results on the NIST dataset}} \\
    \addlinespace[2pt]
    Model & Macro-F1 $\uparrow$ & Micro-F1 $\uparrow$ & EMR $\uparrow$ \\
    \midrule
    UltraIR
    & $\mathbf{0.919_{\pm0.004}}$
    & $\mathbf{0.943_{\pm0.002}}$
    & $\mathbf{0.772_{\pm0.007}}$ \\
    w/o pretraining
    & $0.901_{\pm0.005}$
    & $0.932_{\pm0.002}$
    & $0.719_{\pm0.009}$ \\

    \specialrule{0.8pt}{7pt}{4pt}

    \multicolumn{4}{@{}l@{}}{
    \textbf{b, Molecular structure elucidation}
    \quad \textit{Results on the NIST dataset}} \\
    \addlinespace[2pt]
    Model & Top-1 accuracy $\uparrow$ & Top-5 accuracy $\uparrow$ & Top-10 accuracy $\uparrow$ \\
    \midrule
    UltraIR
    & $\mathbf{0.522_{\pm0.005}}$
    & $\mathbf{0.575_{\pm0.007}}$
    & $\mathbf{0.576_{\pm0.007}}$ \\
    w/o pretraining
    & $0.477_{\pm0.003}$
    & $0.563_{\pm0.006}$
    & $0.567_{\pm0.007}$ \\

    \specialrule{0.8pt}{7pt}{4pt}

    \multicolumn{4}{@{}l@{}}{
    \textbf{c, Mixture-level component quantification}} \\
    \addlinespace[2pt]
    Model
    & Normalized MAE {\scriptsize($\times10^{-3}$)} $\downarrow$
    & Normalized RMSE {\scriptsize($\times10^{-3}$)} $\downarrow$
    & $R^{2}$ $\uparrow$ \\
    \midrule
    UltraIR
    & $\mathbf{2.036_{\pm0.366}}$
    & $\mathbf{2.621_{\pm0.502}}$
    & $\mathbf{0.782_{\pm0.047}}$ \\
    w/o pretraining
    & $2.288_{\pm0.225}$
    & $2.919_{\pm0.370}$
    & $0.714_{\pm0.029}$ \\

    \specialrule{0.8pt}{7pt}{4pt}

    \multicolumn{4}{@{}l@{}}{
    \textbf{d, Soil property prediction}} \\
    \addlinespace[2pt]
    Model
    & Normalized MAE $\downarrow$
    & Normalized RMSE $\downarrow$
    & $R^{2}$ $\uparrow$ \\
    \midrule
    UltraIR
    & $\mathbf{0.086_{\pm0.002}}$
    & $\mathbf{0.325_{\pm0.147}}$
    & $\mathbf{0.921_{\pm0.026}}$ \\
    w/o pretraining
    & $0.104_{\pm0.002}$
    & $0.360_{\pm0.138}$
    & $0.901_{\pm0.025}$ \\

    \bottomrule
    \bottomrule
\end{tabular}

\end{table}

\newpage
\subsection*{Additional functional-group prediction results}

\begin{figure}[!htb]
    \centering
    \includegraphics[width=0.95\textwidth]{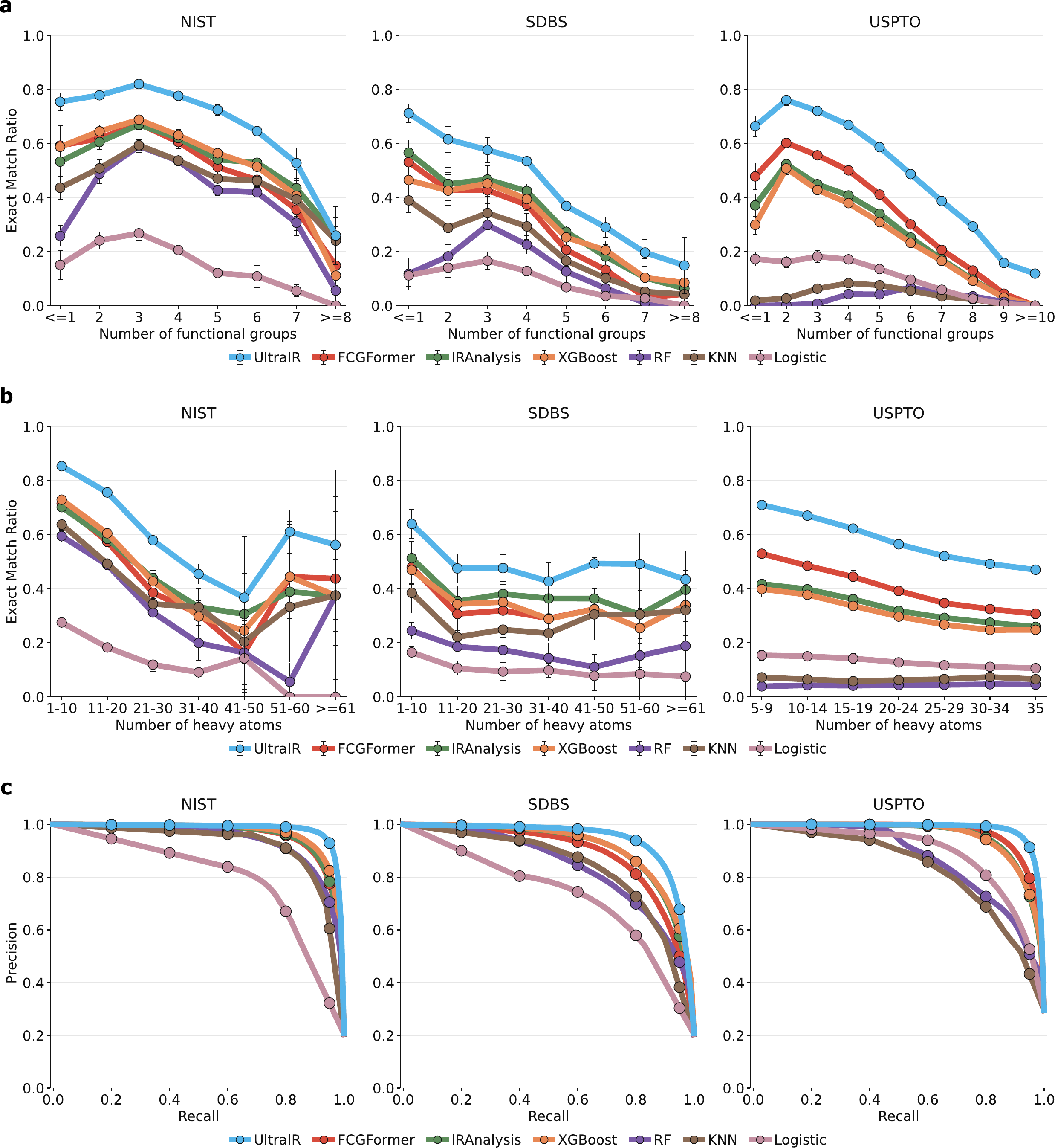}
    \caption{
    \textbf{Functional-group prediction across molecular-complexity strata and precision--recall trade-offs.}
    \textbf{a,} Exact match ratio (EMR) for functional-group prediction stratified by the number of functional groups in each molecule on NIST, SDBS, and USPTO.
    \textbf{b,} EMR for functional-group prediction stratified by the number of heavy atoms in each molecule on NIST, SDBS, and USPTO.
    \textbf{c,} Precision--recall curves for functional-group prediction on NIST, SDBS, and USPTO.
    }
    \label{fig:add_func}
\end{figure}

  \begin{table*}[!t]
    \centering
    \caption{\textbf{Per-functional-group classification performance on the NIST benchmark.}
    F1 scores are reported as means $\pm$ standard deviations across five folds.
    The best result for each functional group is highlighted in bold.}
    \label{tab:nist-per-functional-group-f1}
    \resizebox{\textwidth}{!}{%
    \begin{tabular}{lcccccccc}
    \toprule
    Functional group & Positive rate (\%) & Logistic & KNN & RF
    & XGBoost & IRAnalysis & FCGFormer & UltraIR \\
    \midrule
    Alkane & $79.76$ & $0.919_{\pm0.005}$ & $0.931_{\pm0.008}$ & $0.952_{\pm0.003}$ & $0.966_{\pm0.003}$ & $0.960_{\pm0.003}$ & $0.962_{\pm0.002}$ & $
    \mathbf{0.978_{\pm0.001}}$ \\
    Methyl & $63.08$ & $0.785_{\pm0.009}$ & $0.865_{\pm0.007}$ & $0.877_{\pm0.010}$ & $0.920_{\pm0.005}$ & $0.910_{\pm0.006}$ & $0.911_{\pm0.006}$ & $
    \mathbf{0.950_{\pm0.002}}$ \\
    Alkene & $13.16$ & $0.449_{\pm0.024}$ & $0.684_{\pm0.024}$ & $0.689_{\pm0.027}$ & $0.783_{\pm0.012}$ & $0.761_{\pm0.011}$ & $0.731_{\pm0.008}$ & $
    \mathbf{0.869_{\pm0.015}}$ \\
    Alkyne & $2.04$ & $0.544_{\pm0.047}$ & $0.821_{\pm0.031}$ & $0.747_{\pm0.057}$ & $0.874_{\pm0.023}$ & $0.874_{\pm0.015}$ & $0.884_{\pm0.035}$ & $
    \mathbf{0.949_{\pm0.022}}$ \\
    Alcohols & $23.84$ & $0.682_{\pm0.037}$ & $0.827_{\pm0.008}$ & $0.850_{\pm0.009}$ & $0.887_{\pm0.003}$ & $0.889_{\pm0.008}$ & $0.889_{\pm0.010}$ & $
    \mathbf{0.928_{\pm0.004}}$ \\
    Amines & $22.74$ & $0.621_{\pm0.044}$ & $0.816_{\pm0.005}$ & $0.812_{\pm0.017}$ & $0.865_{\pm0.011}$ & $0.849_{\pm0.005}$ & $0.850_{\pm0.005}$ & $
    \mathbf{0.918_{\pm0.005}}$ \\
    Nitriles & $3.82$ & $0.291_{\pm0.048}$ & $0.528_{\pm0.057}$ & $0.466_{\pm0.048}$ & $0.795_{\pm0.025}$ & $0.659_{\pm0.015}$ & $0.586_{\pm0.033}$ & $
    \mathbf{0.931_{\pm0.016}}$ \\
    Aromatics & $58.84$ & $0.872_{\pm0.016}$ & $0.924_{\pm0.008}$ & $0.920_{\pm0.011}$ & $0.963_{\pm0.004}$ & $0.953_{\pm0.009}$ & $0.956_{\pm0.007}$ & $
    \mathbf{0.982_{\pm0.005}}$ \\
    Alkyl halides & $25.41$ & $0.542_{\pm0.034}$ & $0.764_{\pm0.009}$ & $0.750_{\pm0.010}$ & $0.810_{\pm0.010}$ & $0.800_{\pm0.004}$ & $0.791_{\pm0.013}$ & $
    \mathbf{0.889_{\pm0.010}}$ \\
    Esters & $11.78$ & $0.732_{\pm0.021}$ & $0.884_{\pm0.014}$ & $0.832_{\pm0.029}$ & $0.907_{\pm0.013}$ & $0.912_{\pm0.013}$ & $0.902_{\pm0.019}$ & $
    \mathbf{0.945_{\pm0.010}}$ \\
    Ketones & $8.96$ & $0.419_{\pm0.045}$ & $0.756_{\pm0.023}$ & $0.677_{\pm0.026}$ & $0.783_{\pm0.029}$ & $0.801_{\pm0.011}$ & $0.785_{\pm0.022}$ & $
    \mathbf{0.894_{\pm0.014}}$ \\
    Aldehydes & $1.97$ & $0.552_{\pm0.050}$ & $0.789_{\pm0.034}$ & $0.827_{\pm0.025}$ & $0.873_{\pm0.021}$ & $0.866_{\pm0.035}$ & $0.891_{\pm0.022}$ & $
    \mathbf{0.944_{\pm0.016}}$ \\
    Carboxylic acids & $6.85$ & $0.706_{\pm0.017}$ & $0.856_{\pm0.014}$ & $0.854_{\pm0.017}$ & $0.899_{\pm0.016}$ & $0.881_{\pm0.009}$ & $0.889_{\pm0.012}$ & $
    \mathbf{0.931_{\pm0.015}}$ \\
    Ether & $13.86$ & $0.580_{\pm0.024}$ & $0.784_{\pm0.023}$ & $0.717_{\pm0.022}$ & $0.816_{\pm0.022}$ & $0.818_{\pm0.024}$ & $0.818_{\pm0.020}$ & $
    \mathbf{0.905_{\pm0.014}}$ \\
    Acyl halides & $0.92$ & $0.559_{\pm0.033}$ & $0.827_{\pm0.068}$ & $0.781_{\pm0.046}$ & $0.843_{\pm0.049}$ & $0.864_{\pm0.037}$ & $0.861_{\pm0.050}$ & $
    \mathbf{0.897_{\pm0.062}}$ \\
    Amides & $6.19$ & $0.443_{\pm0.040}$ & $0.680_{\pm0.026}$ & $0.478_{\pm0.052}$ & $0.685_{\pm0.021}$ & $0.712_{\pm0.010}$ & $0.678_{\pm0.016}$ & $
    \mathbf{0.791_{\pm0.014}}$ \\
    Nitro & $5.63$ & $0.761_{\pm0.018}$ & $0.845_{\pm0.026}$ & $0.811_{\pm0.031}$ & $0.889_{\pm0.027}$ & $0.884_{\pm0.017}$ & $0.867_{\pm0.019}$ & $
    \mathbf{0.919_{\pm0.006}}$ \\
    \bottomrule
    \end{tabular}%
    }
  \end{table*}

  \begin{table*}[!t]
    \centering
    \caption{\textbf{Per-functional-group classification performance on the SDBS benchmark.}
    F1 scores are reported as means $\pm$ standard deviations across five folds.
    The best result for each functional group is highlighted in bold.}
    \label{tab:sdbs-per-functional-group-f1}
    \resizebox{\textwidth}{!}{%
    \begin{tabular}{lcccccccc}
    \toprule
    Functional group & Positive rate (\%) & Logistic & KNN & RF
    & XGBoost & IRAnalysis & FCGFormer & UltraIR \\
    \midrule
    Alkane & $74.19$ & $0.858_{\pm0.015}$ & $0.875_{\pm0.015}$ & $0.866_{\pm0.016}$ & $0.915_{\pm0.011}$ & $0.909_{\pm0.012}$ & $0.904_{\pm0.009}$ & $
    \mathbf{0.941_{\pm0.008}}$ \\
    Methyl & $55.78$ & $0.695_{\pm0.039}$ & $0.767_{\pm0.011}$ & $0.776_{\pm0.011}$ & $0.852_{\pm0.011}$ & $0.833_{\pm0.013}$ & $0.815_{\pm0.005}$ & $
    \mathbf{0.881_{\pm0.008}}$ \\
    Alkene & $12.07$ & $0.401_{\pm0.042}$ & $0.470_{\pm0.073}$ & $0.301_{\pm0.085}$ & $0.565_{\pm0.032}$ & $0.614_{\pm0.068}$ & $0.550_{\pm0.059}$ & $
    \mathbf{0.699_{\pm0.050}}$ \\
    Alkyne & $1.07$ & $0.415_{\pm0.089}$ & $0.536_{\pm0.104}$ & $0.102_{\pm0.071}$ & $0.525_{\pm0.067}$ & $0.662_{\pm0.108}$ & $0.576_{\pm0.090}$ & $
    \mathbf{0.726_{\pm0.061}}$ \\
    Alcohols & $29.72$ & $0.631_{\pm0.060}$ & $0.800_{\pm0.026}$ & $0.773_{\pm0.041}$ & $0.844_{\pm0.027}$ & $0.848_{\pm0.018}$ & $0.827_{\pm0.022}$ & $
    \mathbf{0.893_{\pm0.018}}$ \\
    Amines & $27.35$ & $0.587_{\pm0.105}$ & $0.753_{\pm0.008}$ & $0.703_{\pm0.026}$ & $0.783_{\pm0.015}$ & $0.797_{\pm0.019}$ & $0.785_{\pm0.014}$ & $
    \mathbf{0.872_{\pm0.017}}$ \\
    Nitriles & $2.98$ & $0.586_{\pm0.039}$ & $0.278_{\pm0.092}$ & $0.152_{\pm0.077}$ & $0.800_{\pm0.040}$ & $0.752_{\pm0.062}$ & $0.747_{\pm0.049}$ & $
    \mathbf{0.876_{\pm0.025}}$ \\
    Aromatics & $60.47$ & $0.831_{\pm0.017}$ & $0.865_{\pm0.009}$ & $0.855_{\pm0.050}$ & $0.935_{\pm0.014}$ & $0.935_{\pm0.006}$ & $0.929_{\pm0.004}$ &
    $\mathbf{0.968_{\pm0.004}}$ \\
    Alkyl halides & $18.62$ & $0.363_{\pm0.035}$ & $0.429_{\pm0.055}$ & $0.296_{\pm0.045}$ & $0.497_{\pm0.030}$ & $0.557_{\pm0.028}$ &
    $0.487_{\pm0.037}$ & $\mathbf{0.607_{\pm0.035}}$ \\
    Esters & $11.95$ & $0.735_{\pm0.009}$ & $0.799_{\pm0.017}$ & $0.719_{\pm0.024}$ & $0.836_{\pm0.017}$ & $0.852_{\pm0.030}$ & $0.825_{\pm0.026}$ & $
    \mathbf{0.884_{\pm0.018}}$ \\
    Ketones & $9.24$ & $0.384_{\pm0.051}$ & $0.545_{\pm0.040}$ & $0.226_{\pm0.106}$ & $0.569_{\pm0.036}$ & $0.638_{\pm0.030}$ & $0.603_{\pm0.042}$ & $
    \mathbf{0.720_{\pm0.031}}$ \\
    Aldehydes & $2.01$ & $0.251_{\pm0.060}$ & $0.407_{\pm0.075}$ & $0.175_{\pm0.074}$ & $0.446_{\pm0.082}$ & $0.579_{\pm0.092}$ & $0.479_{\pm0.086}$ & $
    \mathbf{0.647_{\pm0.073}}$ \\
    Carboxylic acids & $12.17$ & $0.677_{\pm0.017}$ & $0.791_{\pm0.043}$ & $0.785_{\pm0.039}$ & $0.844_{\pm0.036}$ & $0.846_{\pm0.037}$ &
    $0.827_{\pm0.035}$ & $\mathbf{0.882_{\pm0.030}}$ \\
    Ether & $14.63$ & $0.518_{\pm0.033}$ & $0.615_{\pm0.031}$ & $0.368_{\pm0.092}$ & $0.673_{\pm0.030}$ & $0.709_{\pm0.016}$ & $0.667_{\pm0.025}$ & $
    \mathbf{0.784_{\pm0.021}}$ \\
    Acyl halides & $0.99$ & $0.606_{\pm0.108}$ & $0.728_{\pm0.126}$ & $0.560_{\pm0.157}$ & $0.692_{\pm0.119}$ & $\mathbf{0.740_{\pm0.153}}$ &
    $0.704_{\pm0.098}$ & $0.719_{\pm0.120}$ \\
    Amides & $7.95$ & $0.468_{\pm0.045}$ & $0.612_{\pm0.036}$ & $0.279_{\pm0.111}$ & $0.583_{\pm0.055}$ & $0.672_{\pm0.022}$ & $0.618_{\pm0.022}$ & $
    \mathbf{0.746_{\pm0.030}}$ \\
    Nitro & $6.17$ & $0.685_{\pm0.076}$ & $0.699_{\pm0.066}$ & $0.606_{\pm0.067}$ & $0.794_{\pm0.031}$ & $0.824_{\pm0.025}$ & $0.752_{\pm0.051}$ & $
    \mathbf{0.871_{\pm0.040}}$ \\
    \bottomrule
    \end{tabular}%
    }
  \end{table*}

\begin{table*}[!t]
    \centering
    \caption{\textbf{Per-functional-group classification performance on the USPTO benchmark.}
    F1 scores are reported as means $\pm$ standard deviations across five folds.
    The best result for each functional group is highlighted in bold.}
    \label{tab:uspto-per-functional-group-f1}
    \resizebox{\textwidth}{!}{%
    \begin{tabular}{lcccccccc}
    \toprule
    Functional group & Positive rate (\%) & Logistic & KNN & RF
    & XGBoost & IRAnalysis & FCGFormer & UltraIR \\
    \midrule
    Alkane & $93.15$ & $0.989_{\pm0.001}$ & $0.964_{\pm0.001}$ & $0.965_{\pm0.001}$ & $0.994_{\pm0.000}$ & $0.993_{\pm0.001}$ & $0.997_{\pm0.000}$ & $
    \mathbf{0.999_{\pm0.000}}$ \\
    Methyl & $73.10$ & $0.914_{\pm0.004}$ & $0.839_{\pm0.004}$ & $0.851_{\pm0.004}$ & $0.964_{\pm0.001}$ & $0.964_{\pm0.002}$ & $0.967_{\pm0.001}$ & $
    \mathbf{0.981_{\pm0.001}}$ \\
    Alkene & $11.34$ & $0.160_{\pm0.040}$ & $0.204_{\pm0.006}$ & $0.000_{\pm0.000}$ & $0.240_{\pm0.012}$ & $0.308_{\pm0.018}$ & $0.375_{\pm0.026}$ & $
    \mathbf{0.603_{\pm0.017}}$ \\
    Alkyne & $1.91$ & $0.301_{\pm0.022}$ & $0.165_{\pm0.014}$ & $0.000_{\pm0.000}$ & $0.653_{\pm0.035}$ & $0.644_{\pm0.040}$ & $0.806_{\pm0.020}$ & $
    \mathbf{0.929_{\pm0.013}}$ \\
    Alcohols & $26.22$ & $0.636_{\pm0.018}$ & $0.467_{\pm0.012}$ & $0.634_{\pm0.009}$ & $0.840_{\pm0.005}$ & $0.849_{\pm0.003}$ & $0.905_{\pm0.004}$ & $
    \mathbf{0.953_{\pm0.002}}$ \\
    Amines & $44.48$ & $0.682_{\pm0.022}$ & $0.613_{\pm0.003}$ & $0.659_{\pm0.008}$ & $0.806_{\pm0.004}$ & $0.817_{\pm0.002}$ & $0.857_{\pm0.003}$ & $
    \mathbf{0.917_{\pm0.003}}$ \\
    Nitriles & $6.69$ & $0.426_{\pm0.031}$ & $0.321_{\pm0.015}$ & $0.001_{\pm0.001}$ & $0.813_{\pm0.012}$ & $0.804_{\pm0.014}$ & $0.841_{\pm0.009}$ & $
    \mathbf{0.945_{\pm0.006}}$ \\
    Aromatics & $91.42$ & $0.976_{\pm0.004}$ & $0.954_{\pm0.007}$ & $0.955_{\pm0.013}$ & $0.983_{\pm0.005}$ & $0.983_{\pm0.004}$ & $0.987_{\pm0.002}$ &
    $\mathbf{0.993_{\pm0.001}}$ \\
    Alkyl halides & $46.50$ & $0.653_{\pm0.017}$ & $0.600_{\pm0.002}$ & $0.661_{\pm0.006}$ & $0.731_{\pm0.005}$ & $0.722_{\pm0.004}$ &
    $0.740_{\pm0.003}$ & $\mathbf{0.809_{\pm0.004}}$ \\
    Esters & $18.23$ & $0.650_{\pm0.025}$ & $0.551_{\pm0.024}$ & $0.312_{\pm0.012}$ & $0.790_{\pm0.016}$ & $0.813_{\pm0.015}$ & $0.854_{\pm0.010}$ & $
    \mathbf{0.925_{\pm0.006}}$ \\
    Ketones & $8.22$ & $0.319_{\pm0.041}$ & $0.268_{\pm0.012}$ & $0.002_{\pm0.001}$ & $0.433_{\pm0.043}$ & $0.459_{\pm0.033}$ & $0.550_{\pm0.032}$ & $
    \mathbf{0.692_{\pm0.027}}$ \\
    Aldehydes & $2.67$ & $0.653_{\pm0.037}$ & $0.426_{\pm0.025}$ & $0.117_{\pm0.020}$ & $0.956_{\pm0.009}$ & $0.968_{\pm0.007}$ & $0.967_{\pm0.006}$ & $
    \mathbf{0.988_{\pm0.003}}$ \\
    Carboxylic acids & $9.73$ & $0.806_{\pm0.014}$ & $0.524_{\pm0.019}$ & $0.739_{\pm0.021}$ & $0.947_{\pm0.005}$ & $0.943_{\pm0.003}$ &
    $0.964_{\pm0.002}$ & $\mathbf{0.985_{\pm0.001}}$ \\
    Ether & $37.05$ & $0.656_{\pm0.023}$ & $0.571_{\pm0.015}$ & $0.627_{\pm0.013}$ & $0.785_{\pm0.008}$ & $0.797_{\pm0.007}$ & $0.837_{\pm0.004}$ & $
    \mathbf{0.905_{\pm0.003}}$ \\
    Acyl halides & $0.59$ & $0.132_{\pm0.030}$ & $0.277_{\pm0.044}$ & $0.000_{\pm0.000}$ & $0.205_{\pm0.029}$ & $0.436_{\pm0.049}$ & $0.549_{\pm0.031}$
    & $\mathbf{0.719_{\pm0.026}}$ \\
    Amides & $25.61$ & $0.657_{\pm0.009}$ & $0.620_{\pm0.003}$ & $0.558_{\pm0.015}$ & $0.764_{\pm0.004}$ & $0.796_{\pm0.003}$ & $0.829_{\pm0.003}$ & $
    \mathbf{0.899_{\pm0.003}}$ \\
    Nitro & $5.27$ & $0.346_{\pm0.036}$ & $0.190_{\pm0.017}$ & $0.000_{\pm0.000}$ & $0.518_{\pm0.023}$ & $0.527_{\pm0.017}$ & $0.649_{\pm0.011}$ & $
    \mathbf{0.814_{\pm0.014}}$ \\
    \bottomrule
    \end{tabular}%
    }
  \end{table*}

\clearpage
\subsection*{Additional molecular structure elucidation results}

\begin{table}[!htb]
  \centering
  \caption{\textbf{Ablation of IR and molecular-formula conditioning for molecular structure elucidation on the NIST benchmark.} Tanimoto@$k$ denotes the maximum Morgan Tanimoto similarity between the ground-truth structure and the first $k$ candidates. Results are reported as means $\pm$ standard deviations across five folds.}
  \label{tab:nist-mse-ablation}
  \begin{tabular}{lcccccc}
  \toprule
  \multirow{2}{*}{Model}
  & \multicolumn{2}{c}{Top-1}
  & \multicolumn{2}{c}{Top-5}
  & \multicolumn{2}{c}{Top-10} \\
  \cmidrule(lr){2-3}
  \cmidrule(lr){4-5}
  \cmidrule(lr){6-7}
  & Acc (\%) $\uparrow$ & Tanimoto $\uparrow$
  & Acc (\%) $\uparrow$ & Tanimoto $\uparrow$
  & Acc (\%) $\uparrow$ & Tanimoto $\uparrow$ \\
  \midrule
  UltraIR w/o IR
  & $9.71_{\pm0.21}$ & $0.293_{\pm0.002}$
  & $25.54_{\pm0.48}$ & $0.476_{\pm0.003}$
  & $33.17_{\pm0.43}$ & $0.548_{\pm0.003}$ \\

  UltraIR w/o Formula
  & $38.94_{\pm0.26}$ & $0.594_{\pm0.003}$
  & $46.66_{\pm0.56}$ & $0.676_{\pm0.005}$
  & $49.45_{\pm0.46}$ & $0.702_{\pm0.005}$ \\

  UltraIR
  & $\mathbf{52.20_{\pm0.54}}$ & $\mathbf{0.682_{\pm0.004}}$
  & $\mathbf{57.48_{\pm0.71}}$ & $\mathbf{0.742_{\pm0.003}}$
  & $\mathbf{57.57_{\pm0.74}}$ & $\mathbf{0.748_{\pm0.003}}$ \\
  \bottomrule
  \end{tabular}
\end{table}

\begin{figure}[!b]
    \centering
    \includegraphics[width=0.95\textwidth]{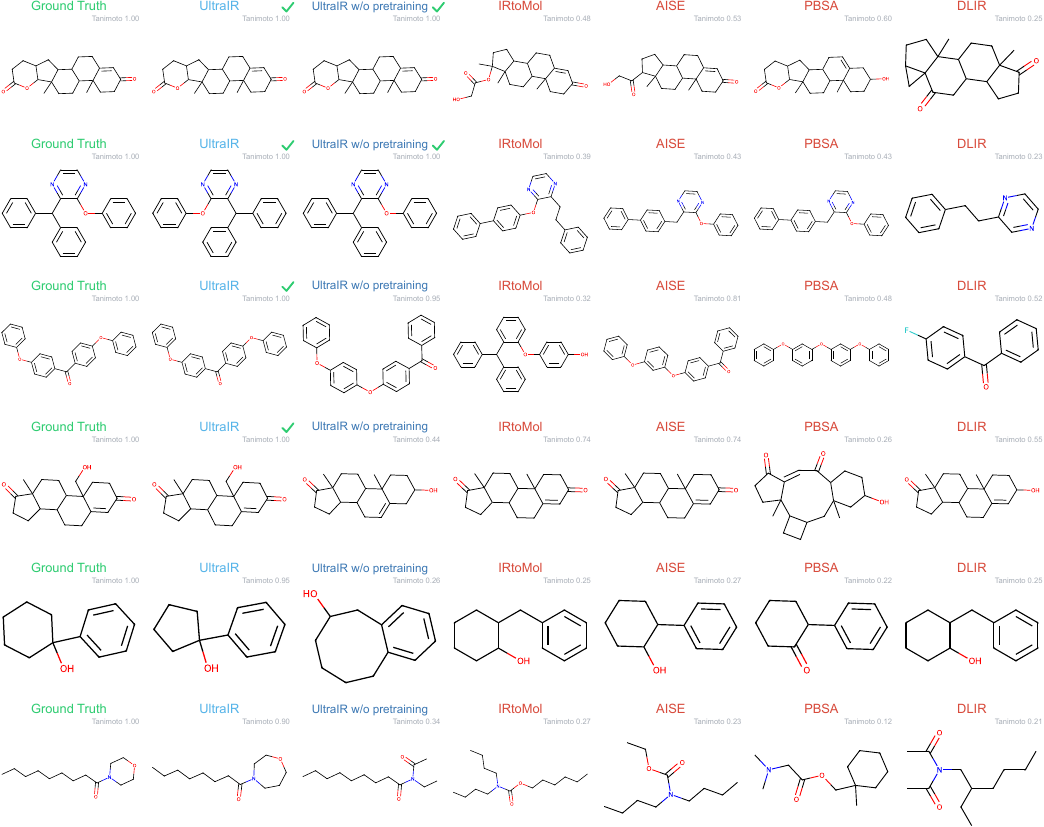}
    \caption{
    \textbf{Additional molecular structure elucidation examples.}
    Six representative examples from the NIST test set comparing the ground-truth molecular structures with candidates generated by UltraIR, UltraIR without pretraining, and the competing structure-elucidation models IRtoMol, AISE, PBSA, and DLIR. Fingerprint-based Tanimoto similarities between each generated candidate and the corresponding ground-truth structure are reported. Green check marks indicate exact structure recovery.
    }
    \label{fig:add_mol}
\end{figure}

\clearpage
\subsection*{Additional physicochemical property prediction results}

\begin{figure}[!htb]
    \centering
    \includegraphics[width=0.95\textwidth]{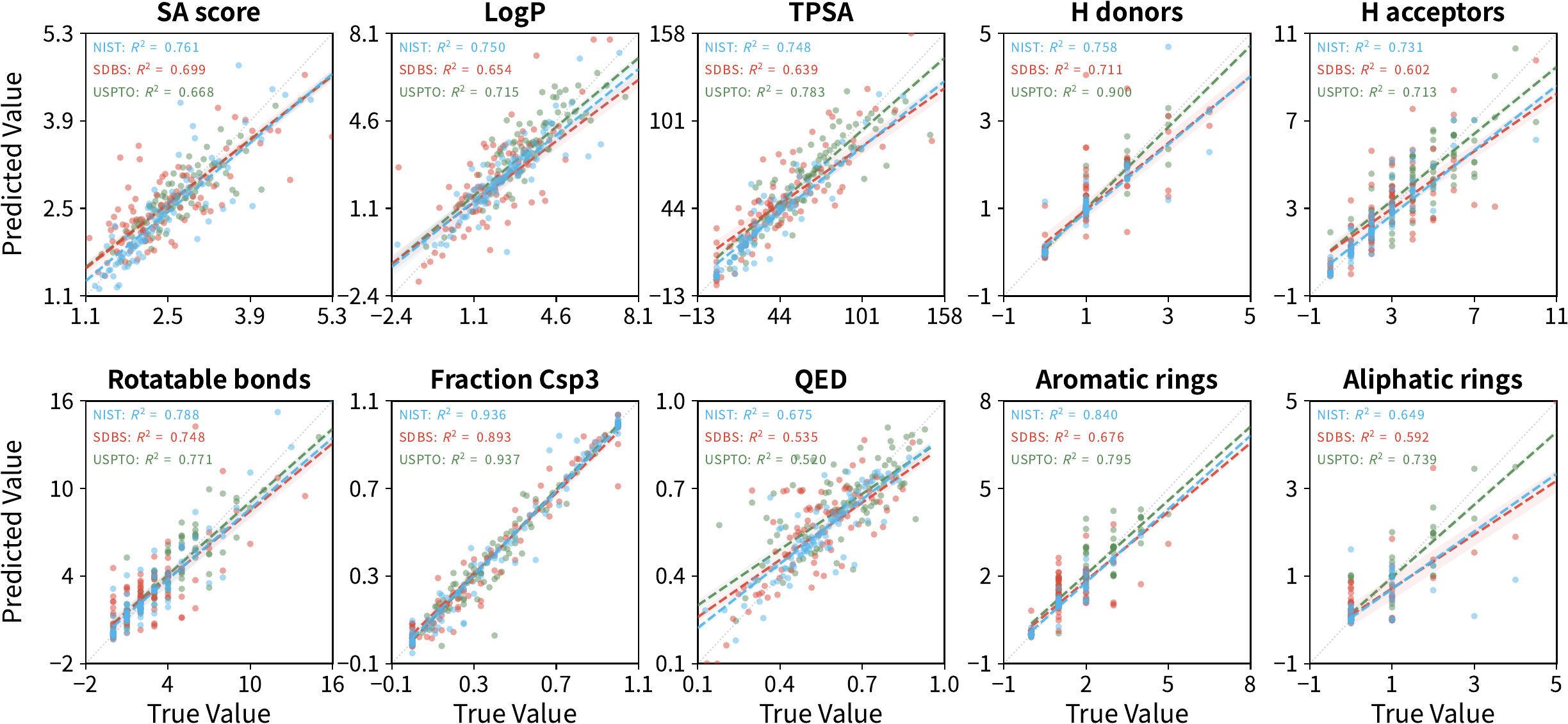}
    \caption{
    \textbf{Property-wise physicochemical prediction performance of UltraIR across NIST, SDBS, and USPTO.}
    Predicted-versus-true parity plots are shown for synthetic accessibility (SA) score, \(\log P\), topological polar surface area (TPSA), hydrogen-bond donors, hydrogen-bond acceptors, rotatable bonds, the fraction of \(sp^3\)-hybridized carbon atoms (Fraction Csp3), quantitative estimate of drug-likeness (QED), aromatic rings, and aliphatic rings. The identity line denotes ideal agreement.
    }
    \label{fig:physicochemical_parity}
\end{figure}

\subsection*{Additional bacterial classification results}

\begin{figure}[!htb]
    \centering
    \includegraphics[width=0.95\textwidth]{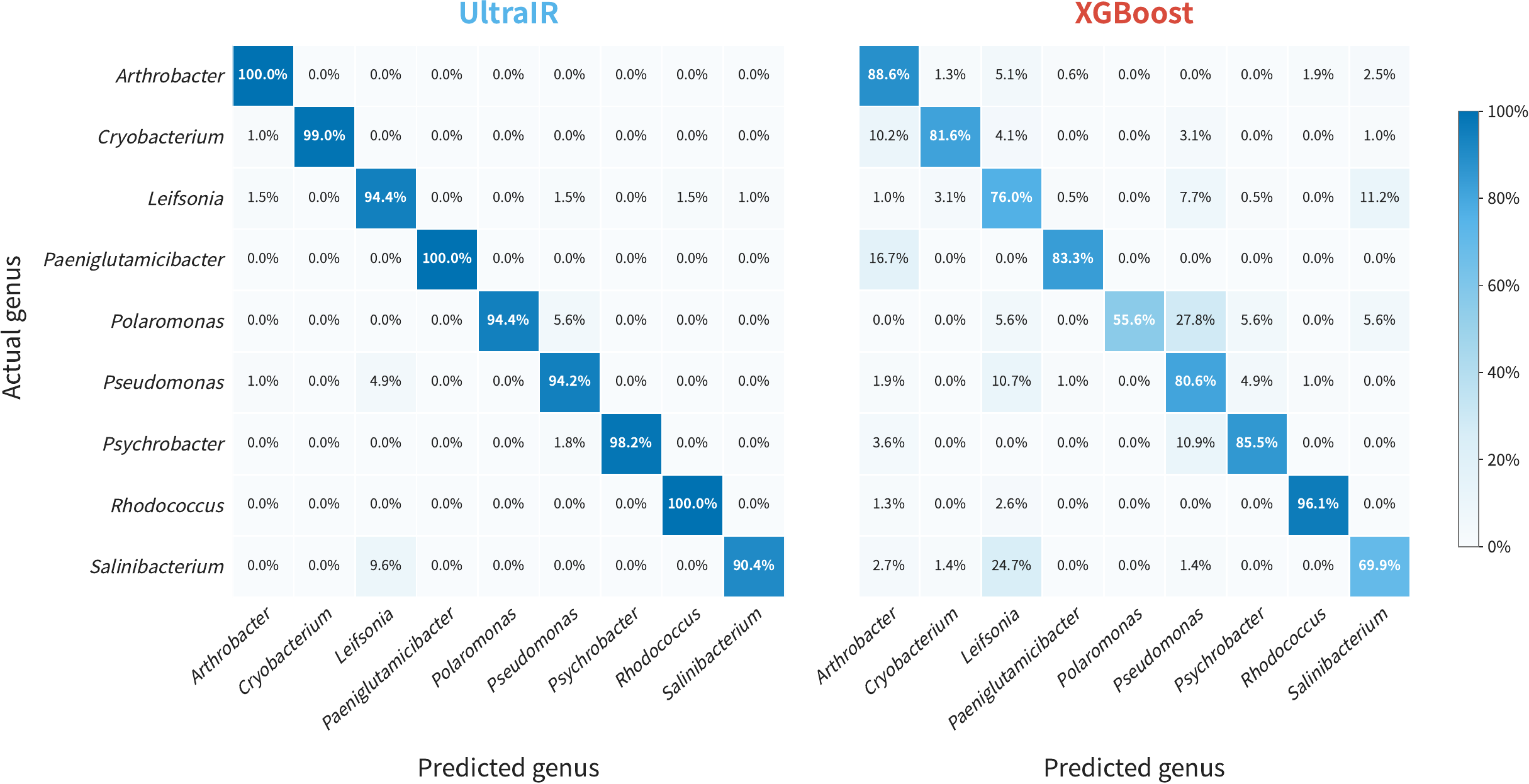}
    \caption{
    \textbf{Confusion matrices for genus-level bacterial classification.}
    Confusion matrices are shown for UltraIR and XGBoost across the nine evaluated bacterial genera. Rows denote actual genera and columns denote predicted genera, with each cell indicating the percentage of samples assigned to the corresponding predicted genus.
    }
    \label{fig:bacterial_confusion}
\end{figure}

\clearpage
\subsection*{Additional medicinal-herb geographic origin traceability results}

\begin{figure}[!htb]
    \centering
    \includegraphics[width=0.95\textwidth]{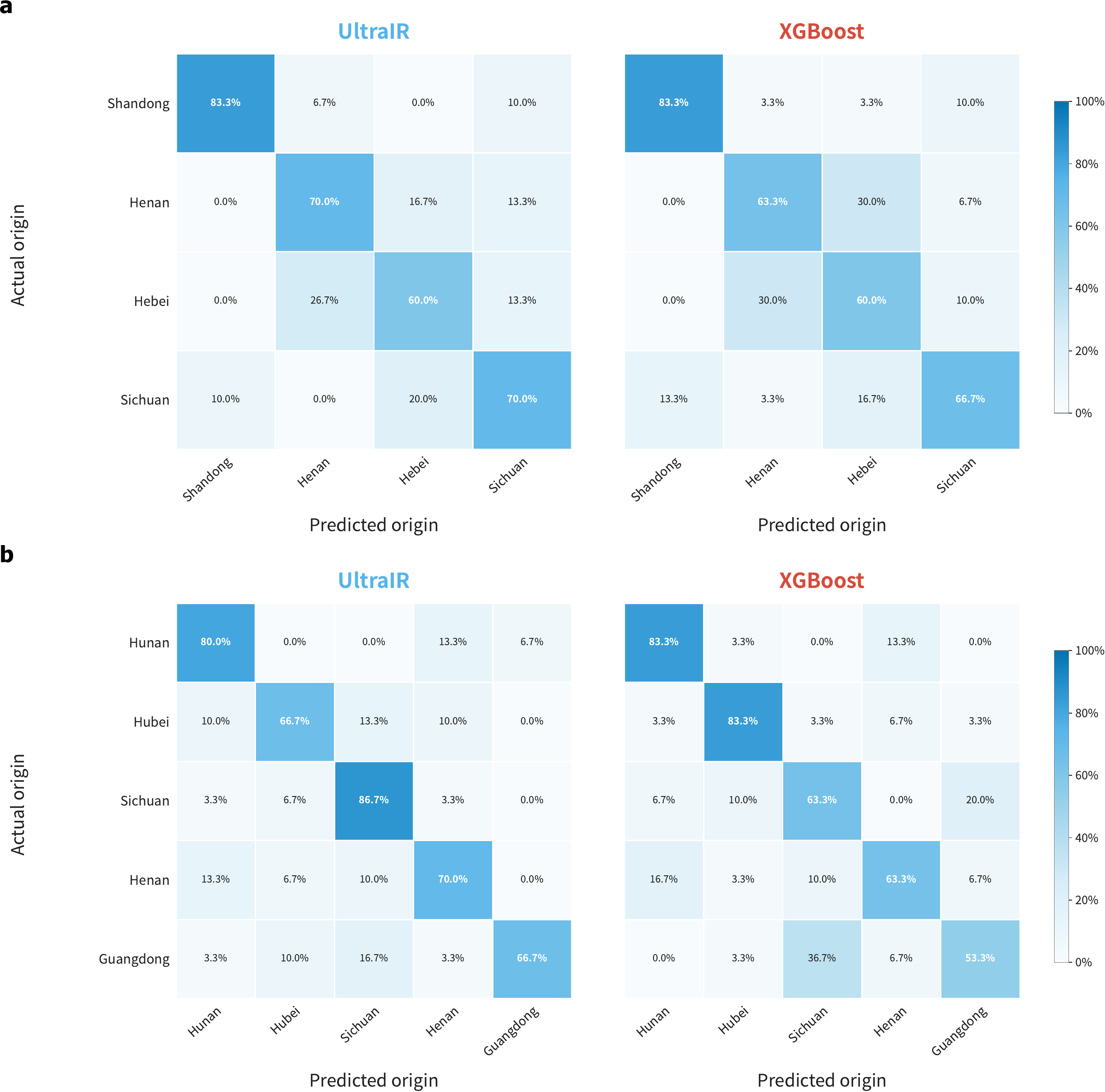}
    \caption{
    \textbf{Confusion matrices for medicinal-herb geographic origin traceability.}
    \textbf{a,} Confusion matrices for Jinyinhua geographic origin traceability, comparing UltraIR and XGBoost across Shandong, Henan, Hebei, and Sichuan.
    \textbf{b,} Confusion matrices for Shanyinhua geographic origin traceability, comparing UltraIR and XGBoost across Hunan, Hubei, Sichuan, Henan, and Guangdong.
    Rows denote actual origins and columns denote predicted origins, with each cell indicating the percentage of samples assigned to the corresponding predicted origin.
    }
    \label{fig:medicinal_herb_confusion}
\end{figure}

\clearpage
\subsection*{Additional microplastics classification results}

\begin{figure}[!htb]
    \centering
    \includegraphics[width=0.95\textwidth]{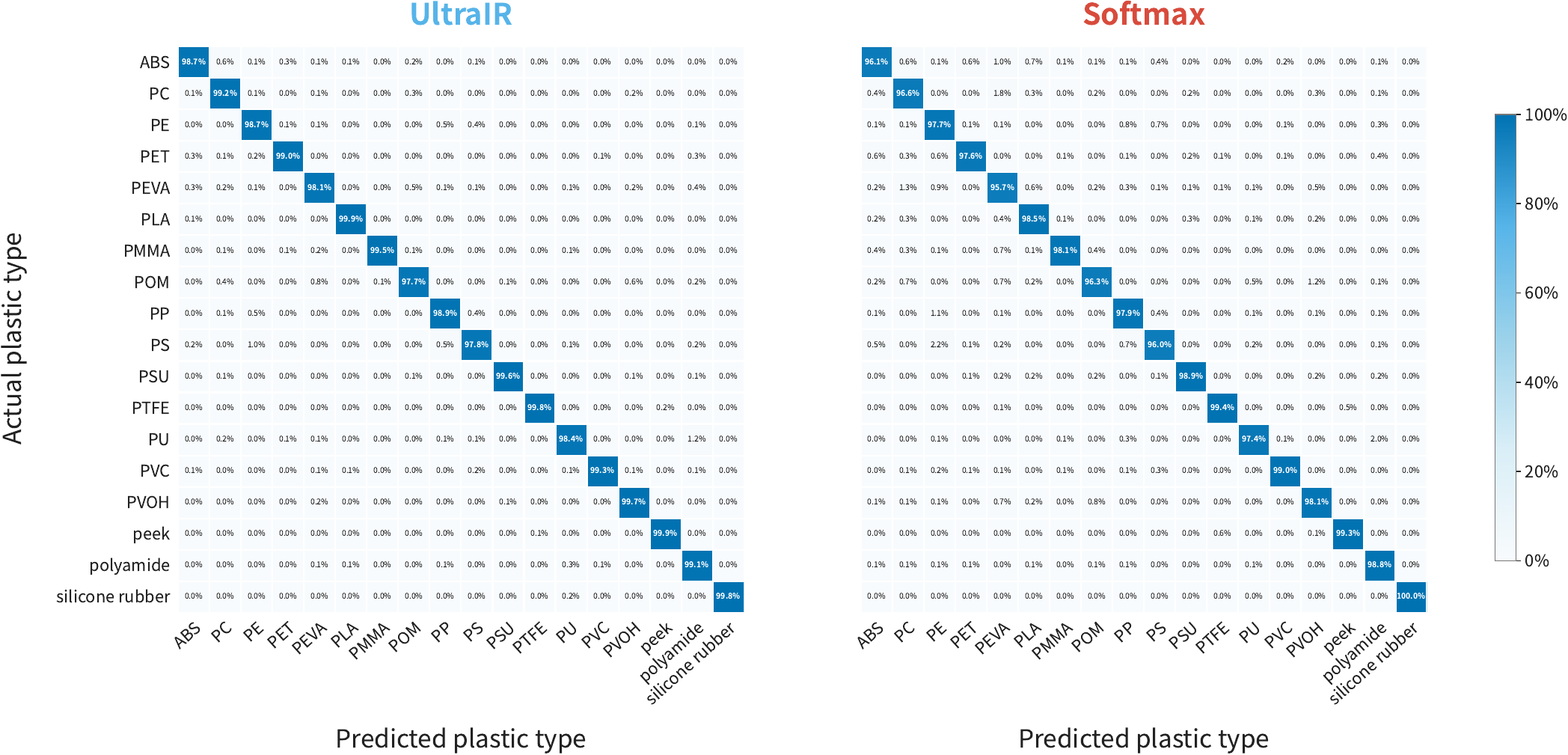}
    \caption{
    \textbf{Confusion matrices for microplastics classification.}
    Confusion matrices are shown for UltraIR and Softmax across the 18 evaluated polymer classes. Rows denote actual plastic types and columns denote predicted plastic types, with each cell indicating the percentage of samples assigned to the corresponding predicted class.
    }
    \label{fig:microplastics_confusion}
\end{figure}

\subsection*{Additional cross-instrument and cross-laboratory generalization results}

\begin{figure}[!htb]
    \centering
    \includegraphics[width=0.95\textwidth]{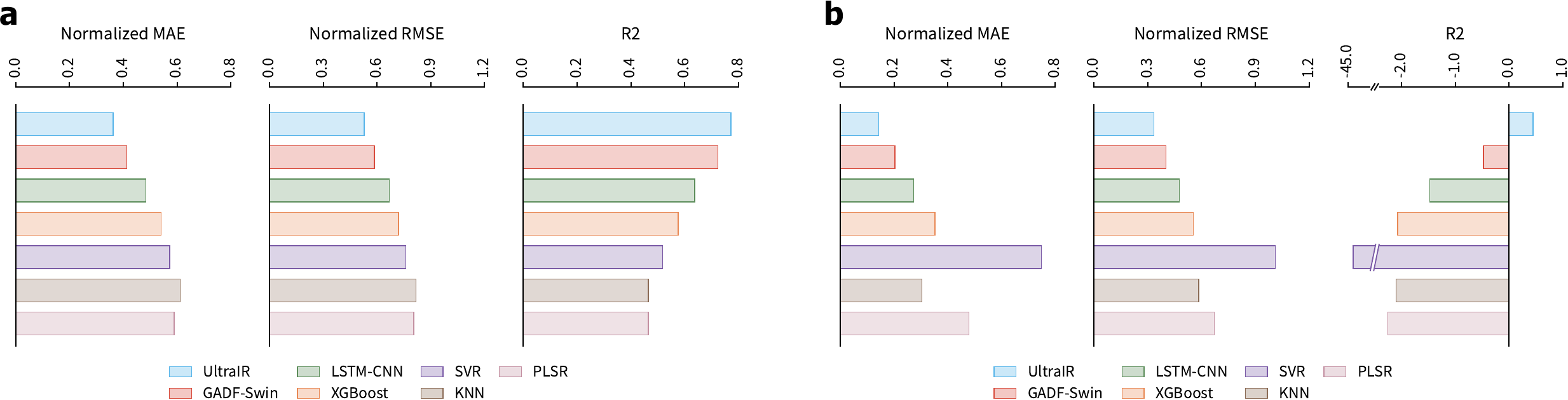}
    \caption{
    \textbf{Cross-instrument and cross-laboratory generalization for soil property prediction.}
    All models were trained on the Kellogg Soil Survey Laboratory (KSSL) subset of OSSL and evaluated by zero-shot inference on the ICRAF--ISRIC subset.
    \textbf{a,} Prediction performance for soil texture and acidity properties, including clay content, silt content, sand content, and pH \((\mathrm{H_2O})\), evaluated using Normalized MAE, Normalized RMSE, and \(R^{2}\).
    \textbf{b,} Prediction performance for soil chemical properties, including cation exchange capacity (CEC) and exchangeable Ca, Mg, K, and Na, evaluated using the same metrics.
    }
    \label{fig:crosslab_soil}
\end{figure}

\end{document}